\documentclass[sigconf]{acmart}

\AtBeginDocument{%
  \providecommand\BibTeX{{%
    \normalfont B\kern-0.5em{\scshape i\kern-0.25em b}\kern-0.8em\TeX}}}

\setcopyright{acmcopyright}
\copyrightyear{2022}
\acmYear{2022}
\acmDOI{10.1145/3535508.3545523}

\acmConference[BCB '22]{13th International ACM Conference on Bioinformatics, Computational Biology and Health Informatics}{August 7--10,
  2022}{Chicago, IL, USA}
\acmPrice{15.00}
\acmISBN{978-1-4503-9386-7/22/08}

\usepackage{makecell}
\usepackage{notoccite}
\usepackage{caption}
\usepackage{subcaption}
\usepackage{booktabs}
\usepackage{multirow}

\begin{document}

\title{Predicting the Need for Blood Transfusion in Intensive Care Units with Reinforcement Learning}

\author{Yuqing Wang}
\email{wang603@ucsb.edu}
\affiliation{%
  \institution{Department of Computer Science, University of California, Santa Barbara}
  \city{Santa Barbara}
  \state{California}
  \country{USA}
}

\author{Yun Zhao}
\email{yunzhao@cs.ucsb.edu}
\affiliation{%
  \institution{Department of Computer Science, University of California, Santa Barbara}
  \city{Santa Barbara}
  \state{California}
  \country{USA}
}

\author{Linda Petzold}
\email{petzold@cs.ucsb.edu}
\affiliation{%
  \institution{Department of Computer Science, University of California, Santa Barbara}
  \city{Santa Barbara}
  \state{California}
  \country{USA}
}

\renewcommand{\shortauthors}{Y. Wang, et al.}

\begin{abstract}
As critically ill patients frequently develop anemia or coagulopathy, transfusion of blood products is a frequent intervention in the Intensive Care Units (ICU). However, inappropriate transfusion decisions made by physicians are often associated with increased risk of complications and higher hospital costs. In this work, we aim to develop a decision support tool that uses available patient information for transfusion decision-making on three common blood products (red blood cells, platelets, and fresh frozen plasma). To this end, we adopt an off-policy batch reinforcement learning (RL) algorithm, namely, discretized Batch Constrained Q-learning, to determine the best action (transfusion or not) given observed patient trajectories. Simultaneously, we consider different state representation approaches and reward design mechanisms to evaluate their impacts on policy learning. Experiments are conducted on two real-world critical care datasets: the MIMIC-III and the UCSF. Results demonstrate that policy recommendations on transfusion achieved comparable matching against true hospital policies via accuracy and weighted importance sampling evaluations on the MIMIC-III dataset. Furthermore, a combination of transfer learning (TL) and RL on the data-scarce UCSF dataset can provide up to $17.02\%$ improvement in terms of accuracy, and up to $18.94\%$ and $21.63\%$ improvement in jump-start and asymptotic performance in terms of weighted importance sampling averaged over three transfusion tasks. Finally, simulations on transfusion decisions suggest that the transferred RL policy could reduce patients' estimated 28-day mortality rate by $2.74\%$ and decreased acuity rate by $1.18\%$ on the UCSF dataset. In short, RL with appropriate patient state encoding and reward designs shows promise in treatment recommendations for blood transfusion and further optimizes the real-time treatment strategies by improving patients' clinical outcomes.
\end{abstract}




\maketitle

\section{Introduction}
In critically ill patients, anemia and coagulopathy are common and associated with poor outcomes, such as increased risk of mortality, myocardial infarction, and thrombosis~\cite{shah2020transfusion}. Transfusion of blood and blood products remains a cornerstone of anemia and coagulopathy treatment in critical care. Clinically, physicians make transfusion decisions mainly based on a patient's hemoglobin level and symptoms of anemia. However, due to urgency in the Intensive Care Units (ICU), physicians may not able to 
comprehensively evaluate all indicators of a patient such as demographics (e.g., age, weight, etc.), medical history (e.g., high blood pressure, diabetes, etc.), and laboratory parameters (e.g., creatinine, hemoglobin, etc.), which can play significant roles in the properness of making a decision about transfusion at a certain time~\cite{liu2021machine}. However, inappropriate decisions on blood transfusion such as the dosage and the type of blood product may even deteriorate the patient status. Thus, developing efficient decision support tools is critical to determining optimal treatment strategies in terms of the need for blood transfusion of each patient and improving the patients' clinical outcomes such as improved survival rates~\cite{zhao2021bertsurv}.

The majority of recent works considered the problem of blood transfusion prediction as a binary classification task~\cite{mitterecker2020machine,walczak2020prediction}, i.e., whether the patient will require a blood transfusion during hospitalization. In practice, however, the transfusion decision that a physician makes at time $t$, based on patients' current situation, can influence the patients' subsequent condition and hence the physician's decision at time $t+1$. Such dynamic status change makes blood transfusion a sequential decision making problem rather than purely a classification task.

In this work, we explore the use of an off-policy batch reinforcement learning (RL) algorithm, namely, the discretized Batch Constrained Q-learning (BCQ)~\cite{fujimoto2019benchmarking} with different state representations and reward functions to provide clinical decision support for the need of blood transfusion for ICU patients. Specifically, we consider transfusion of three common types of blood products: red blood cells (RBC), platelets (PLT), and fresh frozen plasma (FFP). We use two critical care datasets: the Medical Information Mart for Intensive Care III (MIMIC-III)~\cite{johnson2016mimic} and the UCSF. In order to evaluate the performance of the learned policy from sequential patient observations, our experiments are fourfold.

First, we use weighted importance sampling (WIS)~\cite{li2019optimizing} for off-policy evaluation. Second, we compare the RL policy recommendations against the true policies implemented by the hospital regarding whether the patient should receive blood transfusion at each time step using observed patient trajectories along with undertaken interventions. This is based on the assumption that physicians are knowledgeable and experienced to make wise transfusion decisions. Third, we integrate TL to the RL algorithm to improve the original learned policy on the UCSF dataset (target domain) in terms of WIS and accuracy using the knowledge from the MIMIC-III dataset (source domain). Finally, we investigate about how can RL agents assist physicians further optimize real-time treatment strategies on blood transfusion based on the fact that transfusion does not always improve patients' clinical outcomes~\cite{pape2009clinical}. We conduct policy simulations from transferred RL policies to illustrate that blending the RL with what physicians follow could lead to better transfusion strategies and improving patients' short-term (decreased acuity scores) and long-term (decreased mortality) clinical outcomes on the UCSF dataset.

The main contributions of this paper are highlighted as follows:

\begin{enumerate}
    \item[(1)] To the best of our knowledge, this is the first paper to use RL-based approach with different patient state encoding and reward function designs to deal with the blood transfusion policy recommendations in real-world critical care datasets.
    \item[(2)] Experimental results show that TL, together with RL, can improve the transfusion policy learning on the data-scarce UCSF dataset using external knowledge from the MIMIC-III dataset. Specifically, compared to performances without TL, the matching accuracy between the learned policy and the true hospital policy improves up to $17.02\%$. Furthermore, the improvements of jump-start and asymptotic performances in WIS are up to $18.94\%$ and $21.63\%$, respectively.
    \item[(3)] Simulations from transferred RL policies on the UCSF dataset demonstrate both improved short-term and long-term clinical outcomes of ICU patients. Concretely, the overall estimated 28-day mortality rate reduces by $2.74\%$ and the decreased acuity rate during patients' hospital stay reduces by $1.18\%$ compared to the ground truth UCSF transfusion policies.
\end{enumerate}

The remainder of this paper is organized as follows. Section~\ref{related_work} describes related work. The preliminary background is briefly discussed in Section~\ref{prelim}. Section~\ref{datasets} describes the datasets we use for evaluation. The methods we use are outlined in Section~\ref{methods}. Experiments and results are discussed in Section~\ref{results}. Finally, our conclusions are presented in Section~\ref{conclusion}.

\section{Related Work}\label{related_work}
In this section, we review related work on broad applications of RL in healthcare domains, TL approaches and applications in the context of deep RL, as well as existing methods for blood transfusion prediction.

\subsection{RL in healthcare}
RL is popular paradigm for solving sequential decision making problems with sampled, evaluative and delayed feedback simultaneously, and applies broadly in many disciplines, including games~\cite{lample2017playing}, robotics control~\cite{kober2013reinforcement}, and biological data analysis~\cite{mahmud2018applications}. Such distinctive features of RL make it a suitable candidate for developing dynamic treatment regimes (DTRs) that may improve the long-term outcome of patients. For example, cancer treatment, a naturally sequential evolutionary process, is a major objective of RL for DTR application. Various RL techniques have been applied to improve different aspects of cancer treatment~\cite{goldberg2012q}. For other DTR applicatons such as HIV treatment~\cite{ernst2006clinical}, sepsis treatment~\cite{raghu2017deep}, and the need for mechanical ventilation~\cite{prasad2017reinforcement}, we refer readers to~\cite{yu2021reinforcement} for a comprehensive survey of applications of RL techniques in healthcare domains.

\subsection{TL in deep RL}
With the broad prospects of deep RL in different domains, TL has become an important technique to deal with various challenges faced by deep RL, which aims at accelerating the learning process and improving the performance of RL agents by transferring knowledge from external expertise. A significant volume of literature on a wide variety
TL approaches in the context of deep RL focused on different aspects of transferring knowledge such as reward shaping~\cite{ng1999policy}, transfer from demonstrations~\cite{hester2018deep}, policy transfer~\cite{fernandez2006probabilistic}, and inter-task mapping~\cite{taylor2007transfer}.  Recent years have witnessed the remarkable progress that TL combined with deep RL techniques. Such an integration has achieved notable success in applications such as robotics control~\cite{levine2018learning} and game playing~\cite{bellemare2013arcade}. It also demonstrates promising prospects in domains like health informatics~\cite{kosorok2015adaptive} and transportation systems~\cite{wei2018intellilight}.

\subsection{Blood Transfusion Prediction}
In recent years, researchers have exploited the use of machine learning (ML) methods on the problem of blood transfusion prediction. Supervised learning methods such as logistic regression~\cite{kadar2013predicting}, extreme Gradient Boosting (XGBOOST)~\cite{liu2021machine}, random forests~\cite{mitterecker2020machine} or neural networks~\cite{walczak2020prediction}, are applied to predict a binary decision: whether or not a patient will need a transfusion during the hospital stay. Unsupervised switching state autoregressive models on vital signs~\cite{ghassemi2017predicting} are trained to predict whether transfusion was performed at each one-hour interval of the patient’s stay in the hospital. All previous works formulated the blood transfusion prediction as a classification task. In contrast, we propose to use deep RL methods with different state representations of patients and reward functions, in combination with TL techniques, to directly provide sequential treatment recommendations for blood transfusion, and improve ICU patients' clinical outcomes.

\section{Preliminaries}\label{prelim}
In this section, we briefly introduce the typical RL problem formulation via Markov decision process (MDP) and value-based deep RL, and TL from the RL perspective.

\subsection{RL and MDP Formulation}
RL studies sequential decision making processes, generally framed in terms of MDP. A MDP is a 5-tuple ($\mathcal{S}$, $\mathcal{A}, p, r, \gamma$), where each element is defined as follows:
\begin{enumerate}
    \item[(1)] $\mathcal{S}$: a finite state space that the patient is in state $s_t \in \mathcal{S}$ at time $t$.
    \item[(2)] $\mathcal{A}$: a finite action space that the RL agent takes action $a_t \in \mathcal{A}$, which influences the next state $s_{t+1}$.
    \item[(3)] $p(s_{t+1}|s_t, a_t)$: the dynamics of the system, which is the probability of the next state given the current state and action.
    \item[(4)] $r(s_t, a_t, s_{t+1})$: the immediate reward after the transition from $s_t$ to $s_{t+1}$ due to action $a_t$.
    \item[(5)] $\gamma \in [0,1]$: the discount factor, which relates the rewards to the time domain and determines the relative weight in the distant future relative to those in the immediate future.
\end{enumerate}
The purpose of a RL agent is to learn a policy $\pi: \mathcal{S} \rightarrow \mathcal{A}$, i.e., a mapping from a given state $s \in \mathcal{S}$ to a distribution over actions, that maximizes the expected accumulated reward:
\begin{align*}
    R^{\pi} (s_t) = \lim_{T \rightarrow \infty} \mathbb{E}_{s_{t+1}|s_t, \pi} \sum_{t=1}^T \gamma^t r(s_t, a_t, s_{t+1})
\end{align*}
over time horizon $T$. 

\subsection{Value-based Deep RL}
Value-based deep RL methods are used when we use a deep neural network  to approximate the value function. A standard algorithm, deep Q-Network (DQN)~\cite{mnih2015human}, uses a deep convolutional neural network architecture for optimal action-value (known as Q) function approximation. During learning, the Q-learning~\cite{watkins1989learning} update is applied:
\begin{align*}
    \mathcal{L}(\theta) = L_\kappa\bigg(r_{t+1} + \gamma \max_{a_{t+1}} Q_{\theta'}(s_{t+1},a_{t+1}) - Q_{\theta}(s_t, a_t)\bigg),
\end{align*}
where $L_\kappa$ is the Huber loss~\cite{huber1992robust}:
\begin{align*}
    L_\kappa (\delta) = \begin{cases} 
      \frac{1}{2} \delta^2 & \text{if} \ |\delta| \leq \kappa, \\
      \kappa(|\delta| - \frac{1}{2} \kappa) & \text{otherwise}. 
   \end{cases}
\end{align*}
The target network $Q_{\theta'}$ are updated infrequently, where $\theta'$ is updated to $\theta$ after a set number of learning steps. The Huber loss is minimized over replay buffer~\cite{lin1992self}. In healthcare settings, the
dataset is fixed, and there are no further interactions with the environment (here, the patient). Hence, the off-policy batch-mode
deep RL fits naturally.

\subsection{TL in the Context of RL}
Given one or more source domains $\mathcal{M}_s$ and one target domain $\mathcal{M}_t$, TL aims to optimize a policy from $\pi$ to $\pi^\ast$ for the target domain $\mathcal{M}_t$ by leveraging exterior knowledge $\mathcal{D}_s$ from $\mathcal{M}_s$, as well as interior knowledge $\mathcal{D}_t$ from $\mathcal{M}_t$. Here, $\pi = \phi(\mathcal{D}_s \sim \mathcal{M}_s, \mathcal{D}_t \sim \mathcal{M}_t)$, which is a function mapping $\mathcal{S}^t \rightarrow \mathcal{A}^t$ from the states to actions for the target domain $\mathcal{M}_t$. In our problem setting, we have $|\mathcal{M}_s| = |\mathcal{M}_t| = 1$ and knowledge can transfer between two RL agents within the same domain.

\section{Datasets}\label{datasets}
Our patient cohorts are constructed from two datasets: the MIMIC-III (v1.4) and the UCSF dataset. MIMIC-III is a freely available single-center database of critical care data from over 58,000 hospital admissions, including information on 46,520 patients from Beth Israel Deaconess Medical Center between 2001 and 2012. The UCSF dataset, collected from the San Francisco General Hospital and Trauma Center, contains 2,190 highest level trauma activation patients admitted to the Level I trauma center. Both datasets contain de-identified data, including patient demographics, time-stamped measurements from bedside monitoring of vitals, clinical laboratory test results, as well as diagnosis and observations charted by healthcare providers. Both datasets were preprocessed in the same way, including the cohort selection criteria, raw data preprocessing, action space, and reward designs. 


\subsection{Cohort Selection}
From both datasets, we first selected adult patients over the age of 18. Then, patients with less than 24-hour ICU stay or more than 168-hour ICU stay were excluded such that we could focus on patients where transfusion status was likely to impact recovery. After filtering by these criteria, we obtained a final cohort of 15,418 and 2,190 patients for the MIMIC-III and the UCSF datasets, respectively. 
Summary statistics of patient cohorts are summarized in Table~\ref{tab:data_stat}. Here, for each blood transfusion task, patients are considered in two groups: (1) get transfusions at least once during hospitalization (Trans.); (2) do not receive any transfusion during hospitalization (No Trans.). 

\begin{table}[ht]
  \caption{Dataset Statistics}
  \label{tab:data_stat}
  \begin{tabular}{ccc}

    \toprule
    Intervention & \makecell{MIMIC-III \\ Trans. / No Trans.} &  \makecell{UCSF \\ Trans. / No Trans.}\\
    \midrule
    RBC transfusion & 8199 / 7219 & 1572 / 618\\
    PLT transfusion & 1772 / 13646 & 1160 / 1030 \\
    FFP transfusion & 3215 / 12203 & 1358 / 832 \\
  \bottomrule
\end{tabular}
\end{table}

\subsection{Data Preprocessing}
For each patient, we chose vital signs (e.g., heart rate, body temperature, respiration rate) and lab values (e.g., creatinine, hemoglobin, arterial pH) commonly reviewed by clinicians that change over time. Vital signs such as heart rate and temperature are taken several times within an hour, while laboratory tests such as arterial pH and creatinine are administered every few hours as needed. Following~\cite{shung2021neural}, this wide discrepancy in measurement frequency for time-varying continuous features is consolidated into means at 4-hour intervals. A list of clinically reasonable measurement ranges provided by~\cite{harutyunyan2019multitask} is used to remove outlier values for each feature. For the remaining missing values, we applied MICE~\cite{van2011mice} data imputation. After imputation, each feature’s raw data is preprocessed independently by z-scoring across all patients such that the resulting data of each column has zero mean and unit variance. In addition, we extracted some demographic features (e.g., age at admission, admitting weight, gender) for each patient. All demographic features with static values of extracted patient cohorts were fully present. In total, we extracted 42 (4 static and 38 time-varying) features from the MIMIC-III dataset and 38 (9 static and 29 time-varying) features from the UCSF dataset.

\subsection{Action Space}
We define a binary action space for the need of transfusion in a 4-hour window. The action $a_t \in \mathcal{A}$ at each time step is chosen from $a_t \in \{ 0, 1 \}$, which indicates having the patient receiving the transfusion or not (1 indicates presence of transfusion, and 0 indicates absence of transfusion). This discrete action space is suitable for the transfusion of all three blood products (RBC, PLT, and FFP). We choose to use this action space due to the complexity and variations of patients' conditions in clinical practice, and its common definition in existing literature for the blood transfusion prediction task.

\subsection{Reward Design}
Our discrete reward functions are defined in two different ways.
\begin{enumerate}
    \item[(1)]\textit{\textbf{$R1$}}: Since patients' survival is physicians' major objective in critical care, we used the long-term clinical outcome, 28-day mortality status to define the reward. At the terminal time step of each patient’s trajectory, we assign a positive reward $+10$ to patients who survived 28 days after ICU admission and a negative reward $-10$ as a penalty for those who were deceased before 28 days after ICU admission. For all intermediate time steps (including the starting time step), the rewards are all assigned to $0$ since final outcomes of patients are unknown before therapeutic procedures ended. 
    \item[(2)]\textit{\textbf{$R2$}}: During patients' hospitalization, in addition to their final survival, we value short-term outcomes after some treatments by observing an improvement or deterioration of patient status. The acuity score computed at each time step is used to estimate patients' severity of illness and reflect patients' conditions. We use the Sequential Organ Failure Assessment (SOFA) Score~\cite{vincent1996sofa}, a common acuity score, which is suitable to assess both critically ill ICU patients and trauma patients. For the starting time step, we assign the reward $0$ for each patient since the status changed cannot be observed. For all intermediate time steps, the reward function is defined as follows:
    \begin{align*}
         r_{t+1} = \begin{cases} 
      +1 & \text{if} \ s_{t+1}^{SOFA} < s_{t}^{SOFA}, \\
      -1 & \text{if} \ s_{t+1}^{SOFA} > s_{t}^{SOFA}, \\
      0 & \text{otherwise}. 
   \end{cases}
    \end{align*}
This reward function penalizes increasing SOFA scores from $s_t$ to $s_{t+1}$ (deteriorated conditions). If SOFA scores decrease from $s_t$ to $s_{t+1}$, a positive reward is assigned (improved conditions). Otherwise, there is no change in patients' condition, and a reward $0$ will be given. At the terminal time step, we use the 28-day mortality to design the reward, which follows the same way as the 
first reward definition.
\end{enumerate}

\section{Methods}\label{methods}
\begin{figure*}[ht]
\centering
\includegraphics[width=0.9\linewidth]{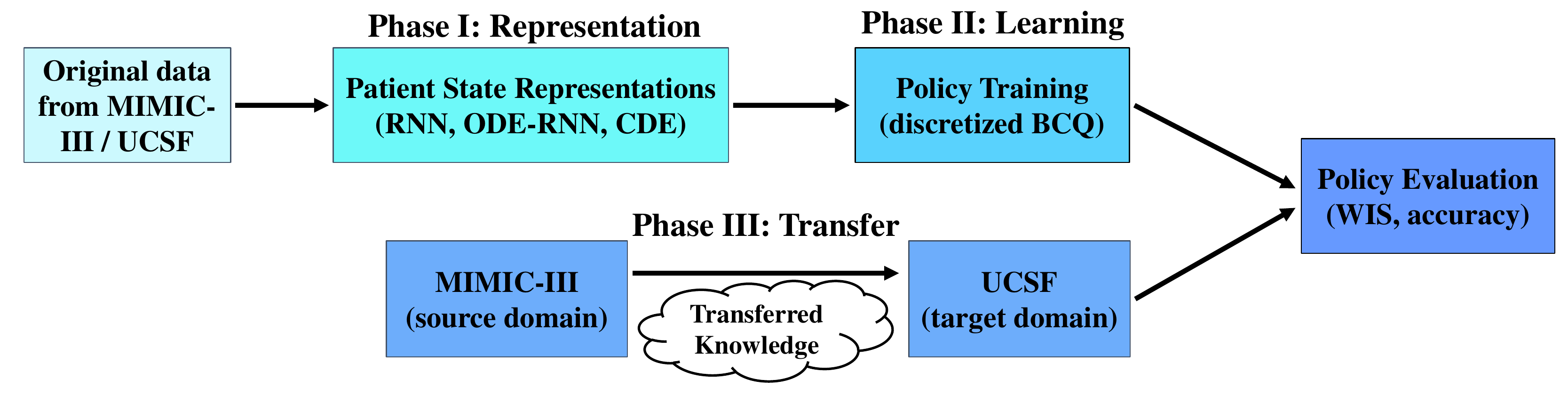}
\caption{An overview of the three-phase framework.}
\label{fig:framework}
\end{figure*}

In this section, we present the overall framework with three phases, including $(I)$ \textbf{representation phase} (patient state representations), $(II)$ \textbf{learning phase} (disretized BCQ), and $(III)$ \textbf{transfer phase} (Q-value transfer and weight transfer from the MIMIC-III to the UCSF). The framework overview is shown in Figure~\ref{fig:framework}.

\subsection{Patient State Representations}\label{state}
\begin{figure}[ht]
     \centering
     \begin{subfigure}[b]{0.25\textwidth}
         \centering
         \includegraphics[width=\textwidth]{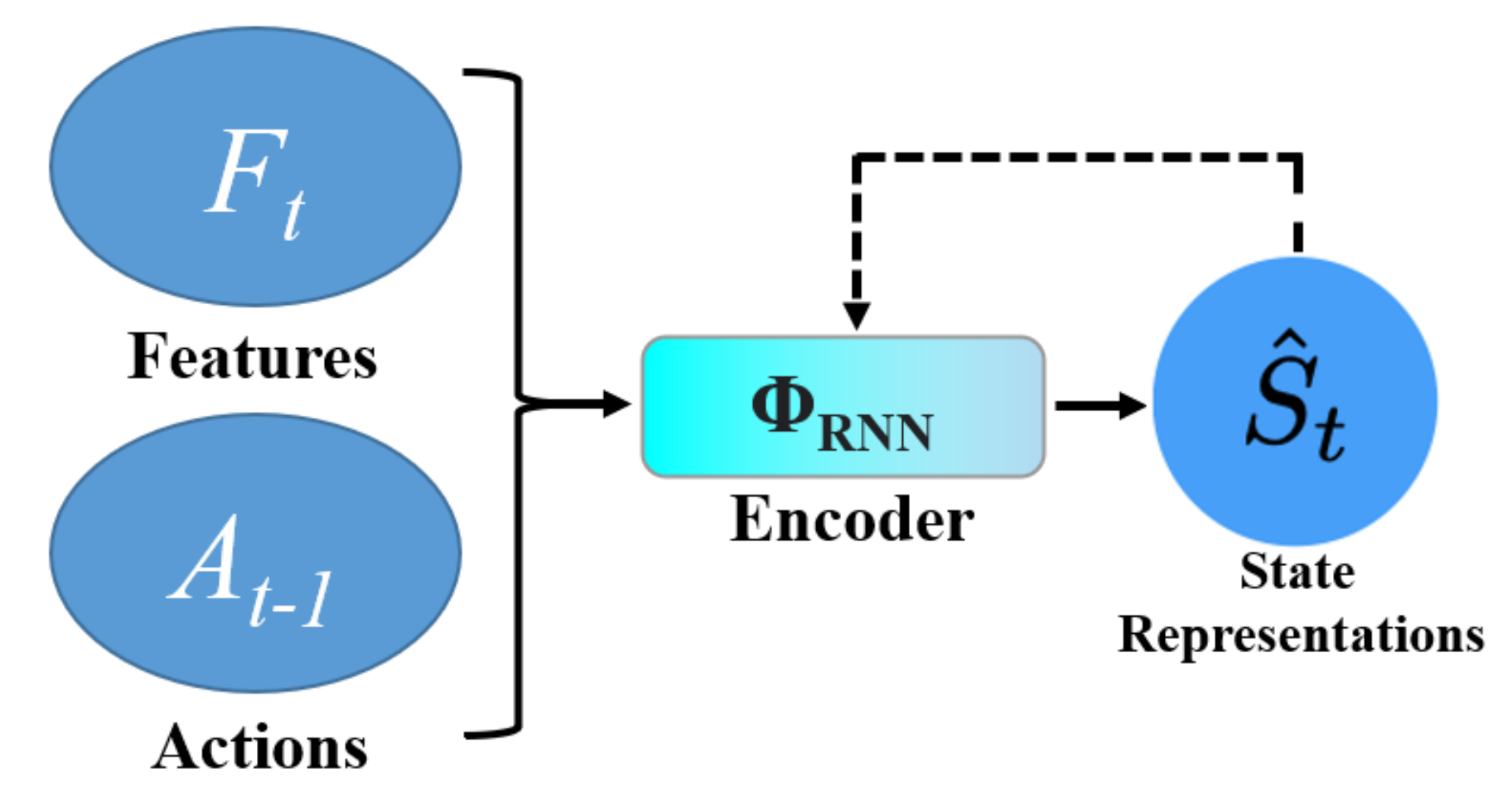}
         \caption{RNN encoder architecture.}
     \end{subfigure}
     \hfill
     \begin{subfigure}[b]{0.25\textwidth}
         \centering
         \includegraphics[width=\textwidth]{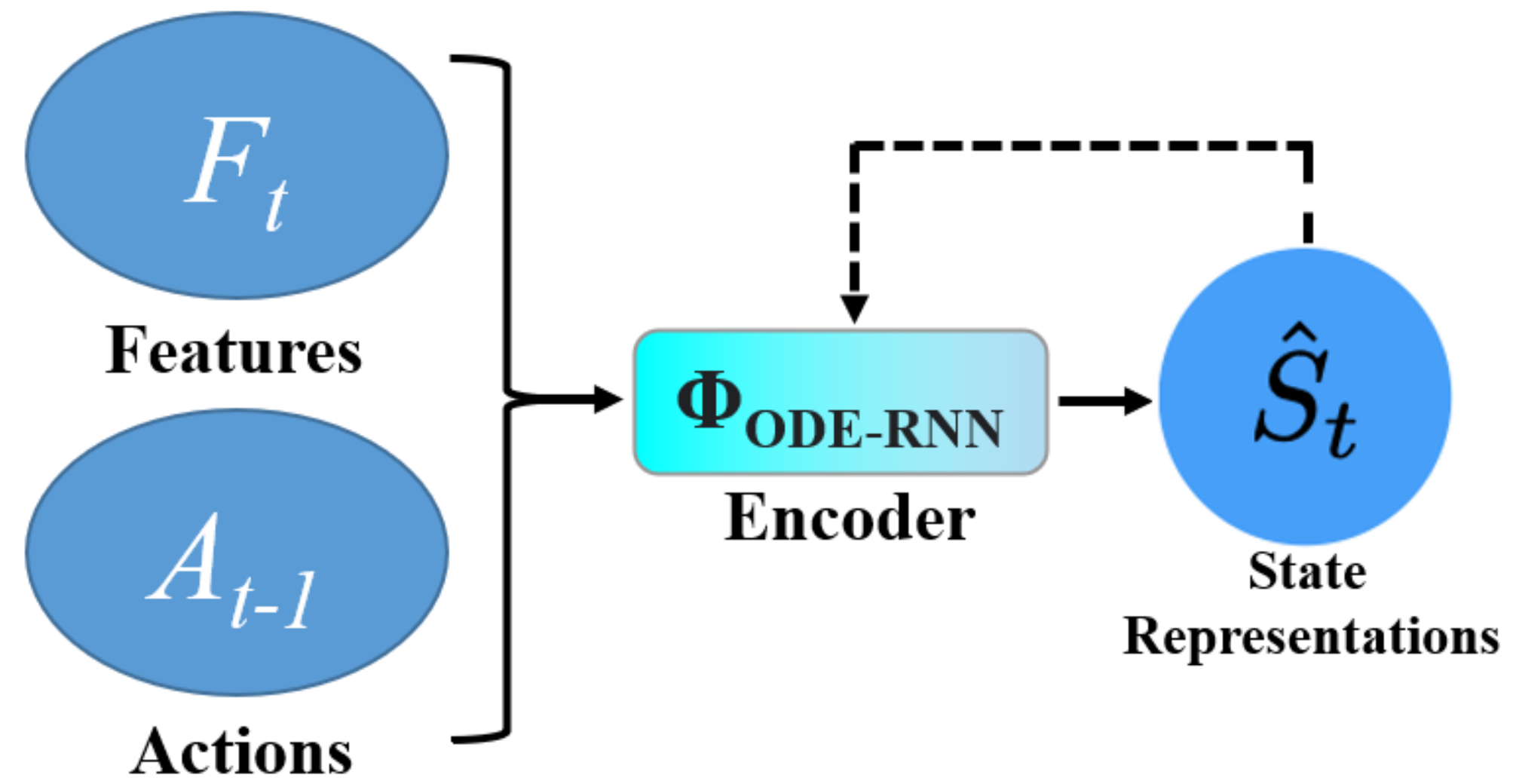}
         \caption{ODE-RNN encoder architecture.}
     \end{subfigure}
     \hfill
     \begin{subfigure}[b]{0.45\textwidth}
         \centering
         \includegraphics[width=\textwidth]{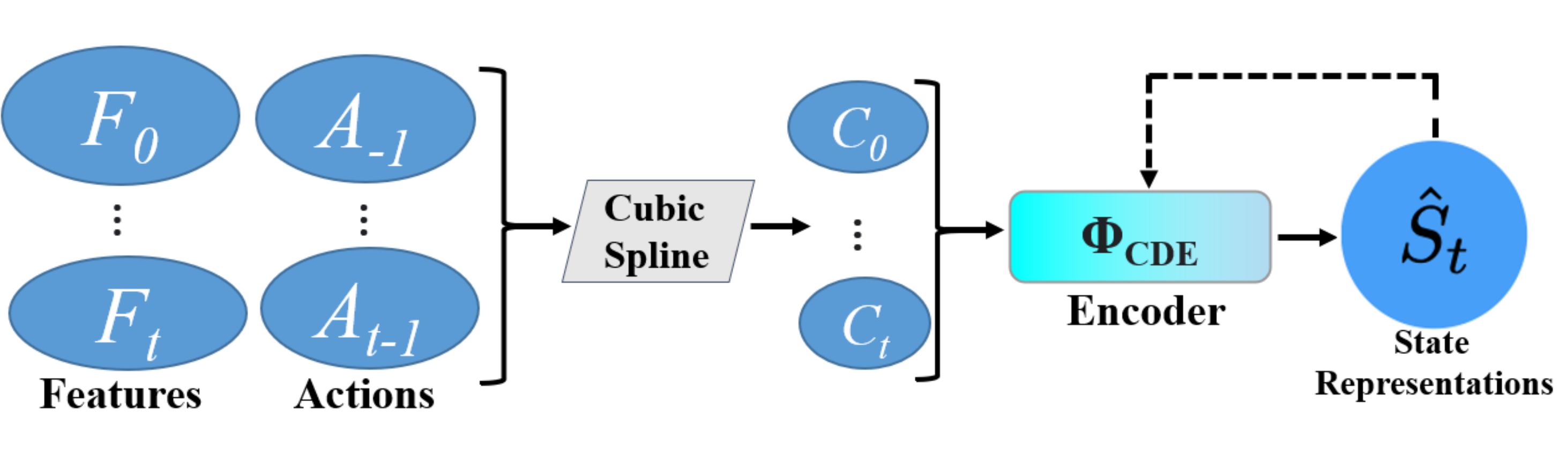}
         \caption{CDE encoder architecture.}
     \end{subfigure}
        \caption{Three recurrent architectures for patient state representations.}
        \label{fig:state_rep}
\end{figure}

We represent patient states via an encoder architecture on both datasets. With a batch of observed patient trajectories, containing transitions between temporal data at time $t$ and $t+1$ with treatment action at time $t$, as well as static demographics, an encoding function $\Phi: \mathcal{F}_{0:t}, \mathcal{A}_{0:{t-1}} \rightarrow \hat{S}_t$ is required to learn the patient state representation. Here, $\mathcal{F}_{0:t}$ represents all feature values from admission to time $t$ and $\mathcal{A}_{0:{t-1}}$ represents actions taken from admission to time $t-1$. Three recurrent architectures are used to represent patients on both datasets, including basic Recurrent Neural Network~\cite{chung2014empirical} (RNN), 
generalized RNN with Ordinary Differential Equations~\cite{rubanova2019latent} (ODE-RNN) and neural Controlled Differential Equations~\cite{kidger2020neural} (CDE). These approach architectures used to construct state representations are depicted in Figure~\ref{fig:state_rep}. \\
\textbf{RNN}: The RNN processes variable-length sequences by
utilizing a recurrent hidden state, which is activated by
features propagated from the previous time step. In our setting, given the current feature value $F_t$, it is concatenated with the previous
action $A_{t-1}$, and passed into the RNN along with the
hidden state representation from the previous
time step $\hat{S}_{t-1}$, resulting in the current hidden state representation $\hat{S}_{t}$.\\
\textbf{ODE-RNN}: An ODE-RNN generalizes RNNs to have continuous-time hidden dynamics defined by ODEs. The main difference between ODE-RNN and basic RNN is that the recurrent hidden state updates based on an ODE between feature observations instead of being fixed. \\
\textbf{CDE}: A neural CDE is the continuous analogue of an RNN. It is similar to ODE-RNNs in terms of the temporal modeling dynamics by parameterizing the time derivative of a hidden state. Different from ODE-RNN, the hidden states in CDEs evolve smoothly as a
function of time. A natural cubic spline interpolation is used to achieve continuous dependency on the data throughout the entire latent trajectory. Then, the network operates on pre-computed cubic spline coefficients instead of real feature values. At $t=0$, the
value for the initial latent space is is computed by a linear map on the inputs.

\subsection{Discretized Batch Constrained Q-learning}
We train RL policies on each of the learned state
representations described in Section~\ref{state}. We seek to learn a policy to select the optimal transfusion action using the state representation: $A_t \sim \pi (\hat{S}_t|\mathcal{F}_{0:t}, \mathcal{A}_{0:{t-1}})$. In our setting, we learn a policy via an off-policy batch RL algorithm, namely, discretized BCQ~\cite{fujimoto2019benchmarking}. This method overcomes the issue of extrapolation errors, which occur in standard off-policy algorithms such as DQN. The discretized BCQ mainly include a state-conditioned model $G_\omega$, a Q-network $Q_\theta$, and a target Q-network $Q_{\theta'}$. The model $G_\omega$ is constructed using behavior cloning, which is trained in the way of supervised learning with cross-entropy loss. The Q-network $Q_\theta$ is updated as follows:
\begin{align*}
    \mathcal{L}(\theta) = L_\kappa \bigg(r_{t+1} + \gamma \max_{a_{t+1}|G_\omega (a_{t+1}|s_{t+1})/ \underset{\hat{a}}{\text{max}} G_\omega (\hat{a}|s_{t+1}) > \tau}Q' - Q\bigg)
\end{align*}
where $Q' = Q_{\theta'}(s_{t+1}, a_{t+1})$ and $Q = Q_{\theta}(s_t, a_t)$. Here, the threshold $\tau$ is used to select actions with higher probability, a constraint for state-action pairs. The final policy learned is based on the greedy behavior of $\tau$:
\begin{align*}
    \pi(s) = \operatorname*{argmax}_{a_t|G_\omega(a_t|s_t) / \underset{\hat{a}}{\text{max}} G_\omega (\hat{a}|s_t) > \tau} Q_\theta (s_t,a_t).
\end{align*}
See~\cite{fujimoto2019benchmarking} for a more detailed algorithm description.

\subsection{Transfer discretized BCQ}\label{BCQ_transfer}
In consideration of the amount of data in our proprietary and public datasets, we explore the use of TL from the MIMIC-III (expert model) to the UCSF (learner model). We consider two ways of transfer: Q-value transfer and weight transfer. \\
\textbf{Q-value Transfer (BCQ-QVT)}: The expert model learns its policies using the discretized BCQ algorithm, and selects actions based on its own Q-values. The learner model uses the Q-values
from the expert model to adjust its network parameters in the
direction that may attain better performance than before and help speed up convergence. The Q-values from the expert model are used in the loss function to guide the learner model, resulting in the following formulation of the loss function:
\begin{align*}
    \mathcal{L}(\theta) = L_\kappa \bigg(r_{t+1} + \gamma \max_{a_{t+1}|G_\omega (a_{t+1}|s_{t+1})/ \underset{\hat{a}}{\text{max}} G_\omega (\hat{a}|s_{t+1}) > \tau}Q' - Q + Q''\bigg)
\end{align*}
where $Q' = Q_{\theta'}(s_{t+1}, a_{t+1})$, $Q = Q_{\theta}(s_t, a_t)$, and $Q'' = Q_{\text{exp},\theta''} (s_t, a_t)$ (Q-values from the expert model with parameter set $\theta''$). \\
\textbf{Weight Transfer (BCQ-W)}: We consider either retraining layers or a combination of retraining and re-initializing layers during transfer. Specifically, retraining layers involves initializing layers with the weights of a pre-trained policy and continuing to update these weights with backpropagation. Re-initializing layers involves randomly initializing the weights for a layer, rather than using the pre-trained weights.

\section{Experiments and Results}\label{results}
Our experiments explore: $(1)$ RL off-policy evaluation via WIS, $(2)$ the matching accuracy between RL policy recommendations and ground truth actions performed by the hospital, $(3)$ a combination of TL and RL from the MIMIC-III to the UCSF, $(4)$ policy simulation from the transferred RL policy on the UCSF dataset. All the reported results and analysis through the remainder of this Section are
provided using only the testing subset of the patient cohort.

\begin{table*}[htbp]
  \caption{Hyperparameter search space for tuning of classification algorithms on both datasets.}
  \label{tab:classification_hyper}
  \begin{tabular}{ccc}

    \toprule
   & Tuning parameters & Search Space\\
    \midrule
    LR & Inverse of regularization strength & $[ 1\mathrm{e}{-3}, 1\mathrm{e}{-2}, 1\mathrm{e}{-1}, 1, 10, 100, 1000]$ \\
    \hline
    RF & \makecell{Number of trees in the forest \\ Maximum depth of the tree} & \makecell{$[25, 50, 75, 100, 125, 150, 175, 200]$ \\ $[2, 5, 8, 10, 12, 15, 20]$} \\
    \hline
    XGBOOST & \makecell{Number of trees in the forest \\ Maximum depth of the tree} & \makecell{$[25, 50, 75, 100, 125, 150, 175, 200\}$ \\ $\{2, 5, 8, 10, 12, 15, 20]$} \\
    \hline
    MLP & \makecell{Hidden layer size \\ Batch size \\ Activation function \\ Optimizer \\ Learning rate} & \makecell{$[16, 32, 64, 128, 256, 512]$ \\ $[8, 16, 32, 64, 128, 256]$ \\ $[\text{ReLU}, \text{tanh}, \text{Sigmoid}]$ \\ $[\text{SGD}, \text{Adam}]$ \\ $[ 1\mathrm{e}{-4}, 1\mathrm{e}{-3}, 1\mathrm{e}{-2}, 1\mathrm{e}{-1}]$} \\
  \bottomrule
\end{tabular}
\end{table*}

\begin{table*}[htbp]
  \caption{Hyperparameter search space for tuning of discrete BCQ algorithm on both datasets.}
  \label{tab:bcq_hyper}
  \begin{tabular}{cc}

    \toprule
    Tuning parameters & Search Space\\
    \midrule
     Number of nodes per layer in Q-network  & [32, 64, 128]\\
     Batch size & [8, 16, 32, 64, 128, 256, 512] \\
     Optimizer & [SGD, Adam] \\
     Discount factor $\gamma$ & [0.97, 0.975, 0.98, 0.985, 0.99, 0.995] \\
     Target Q-network update frequency & [1k, 2k, 4k, 8k training iterations] \\
     Learning rate & [$1\mathrm{e}{-6}, 5\mathrm{e}{-6}, 1\mathrm{e}{-5}, 5\mathrm{e}{-5}, 1\mathrm{e}{-4}, 5\mathrm{e}{-4}, 1\mathrm{e}{-3}, 5\mathrm{e}{-3}$] \\
     Threshold $\tau$ & [0.1, 0.15, 0.2, 0.25, 0.3, 0.35, 0.4, 0.45, 0.5] \\
     Huber loss $\kappa$ & [0.8, 0.9, 1.0, 1.1, 1.2] \\
  \bottomrule
\end{tabular}
\end{table*}

\begin{figure*}[htbp]
    \centering
    {\includegraphics[width=0.33\textwidth,]{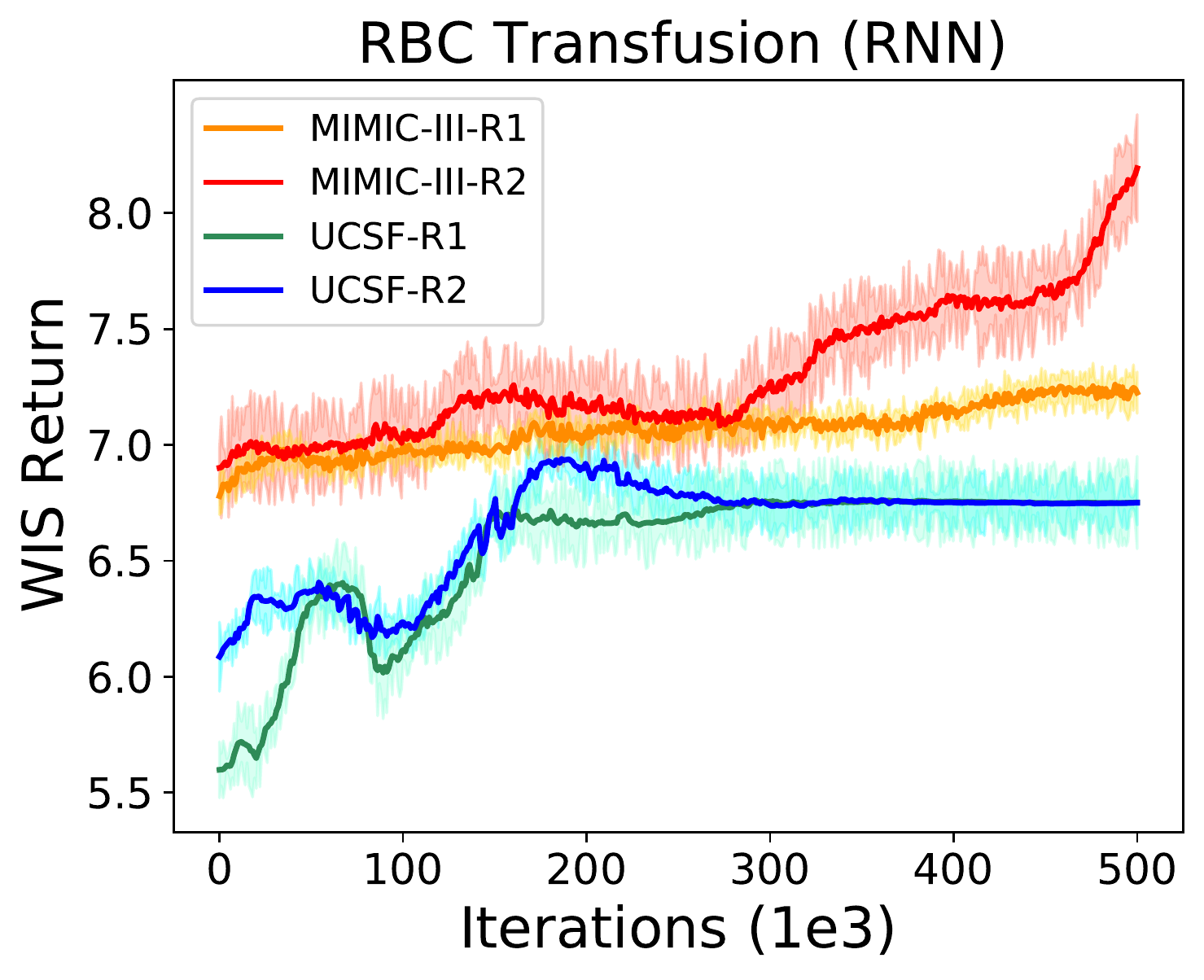}}
    \hfill
    {\includegraphics[width=0.33\textwidth, ]{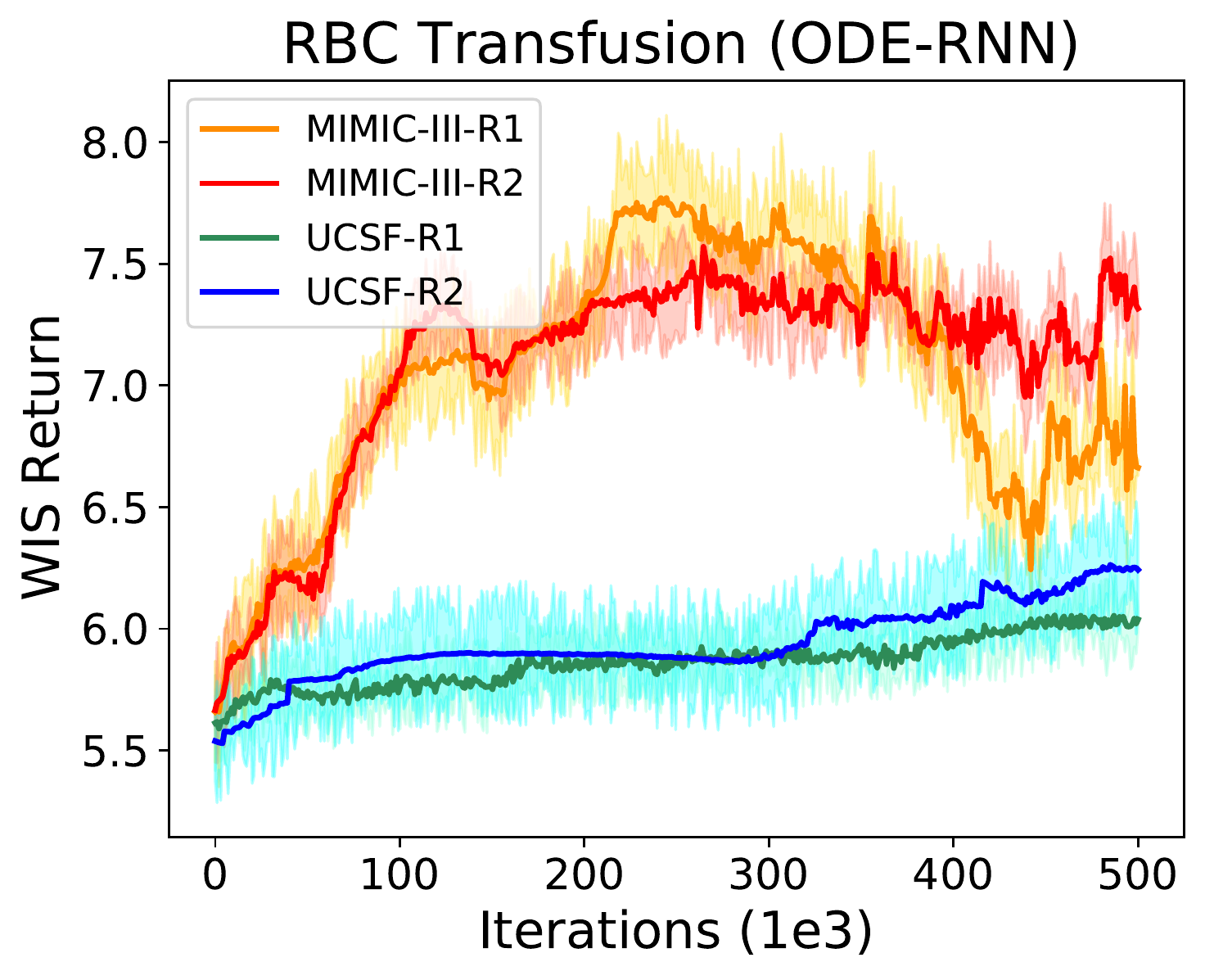}}
    \hfill
    {\includegraphics[width=0.33\textwidth, ]{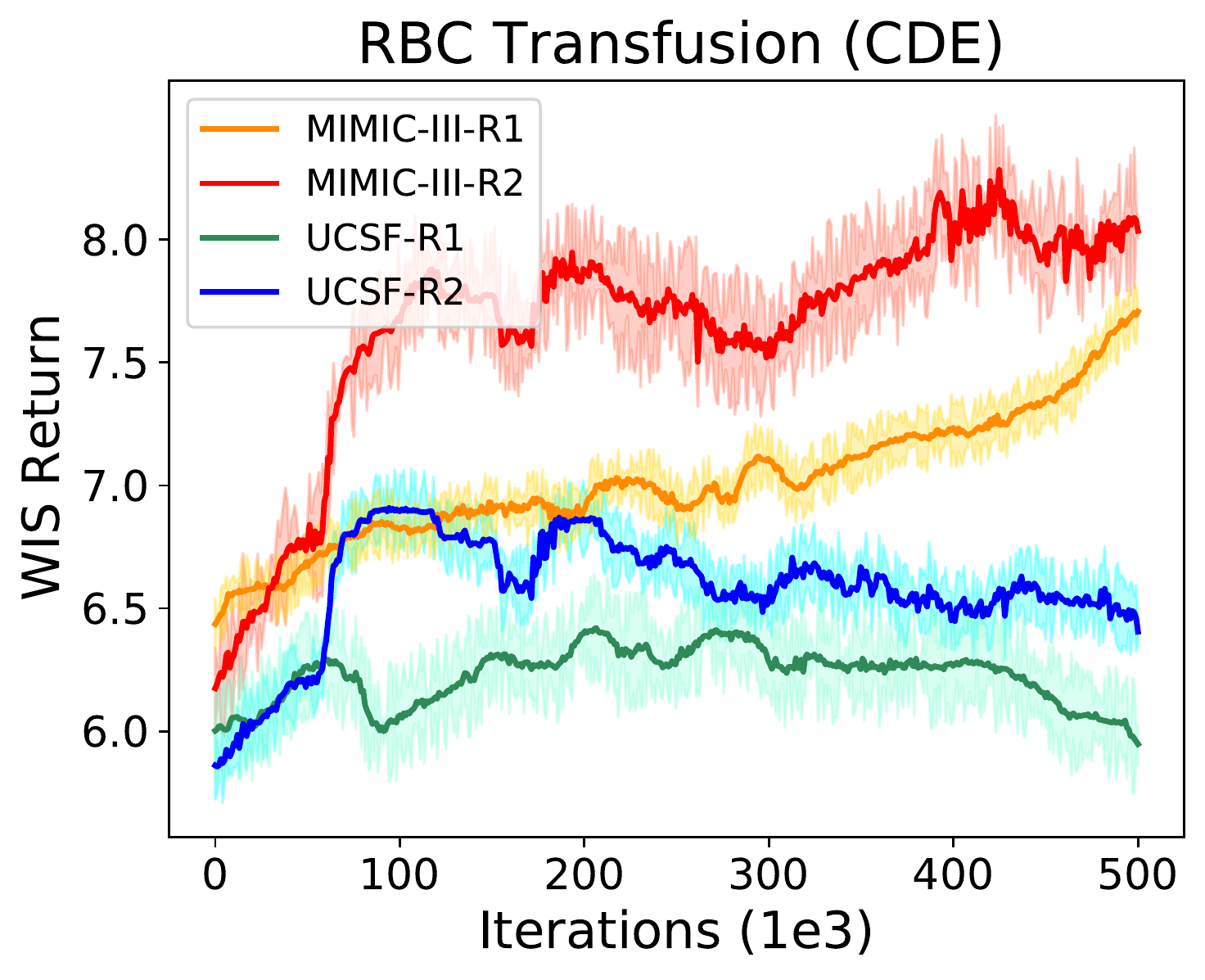}}
    \hfill
    {\includegraphics[width=0.33\textwidth,]{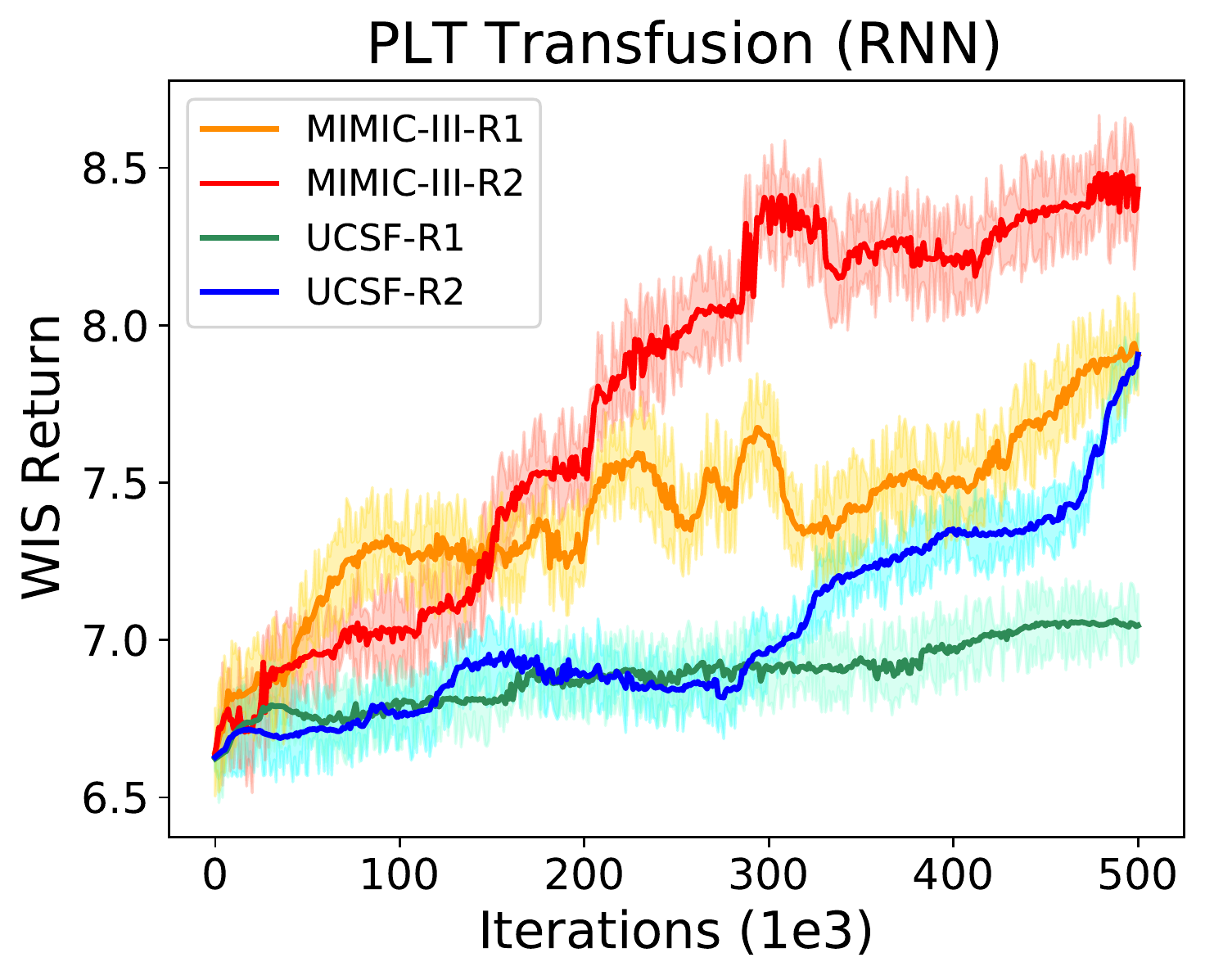}}
    \hfill
    {\includegraphics[width=0.33\textwidth, ]{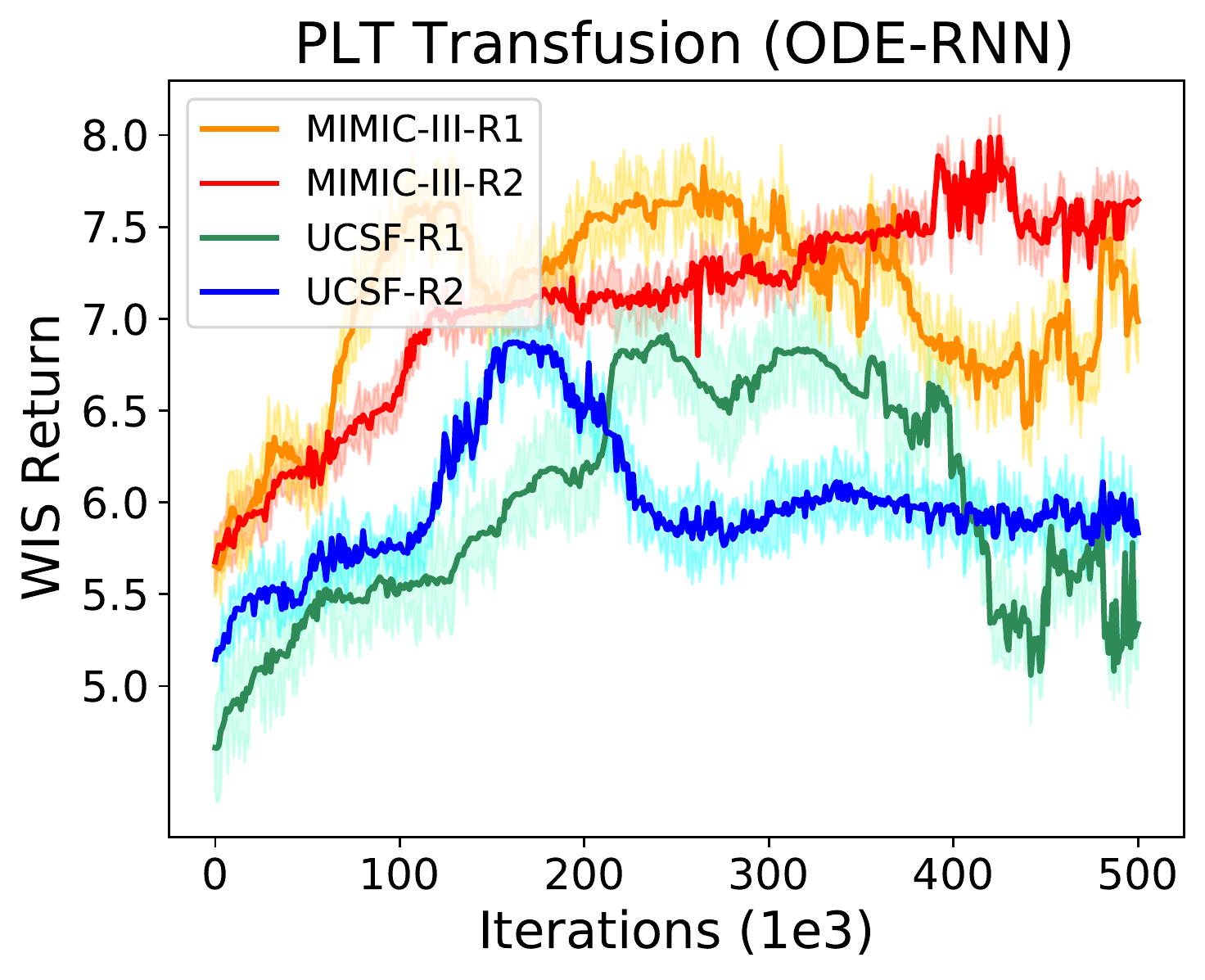}}
    \hfill
    {\includegraphics[width=0.33\textwidth, ]{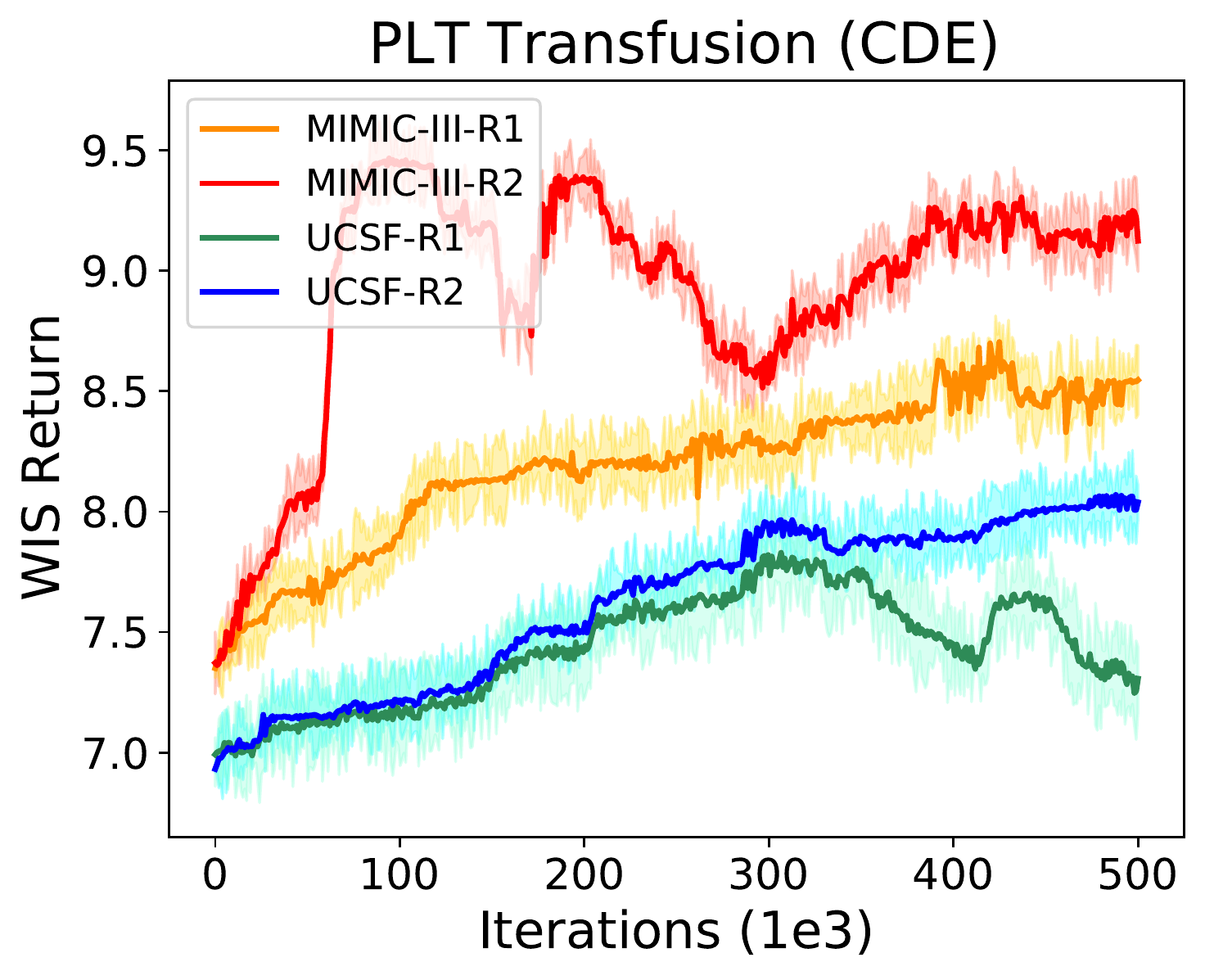}}
    \hfill
    {\includegraphics[width=0.33\textwidth,]{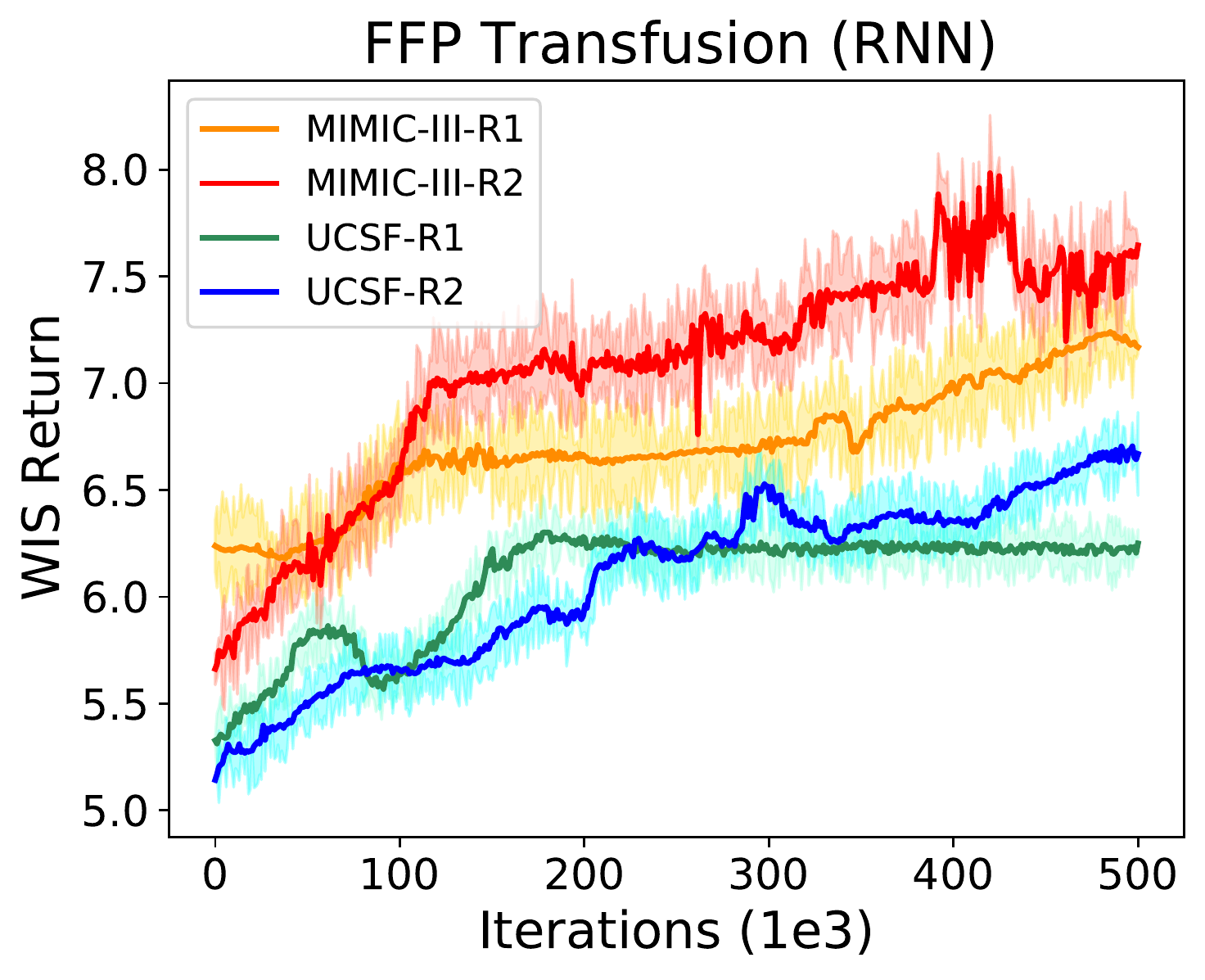}}
    \hfill
    {\includegraphics[width=0.33\textwidth, ]{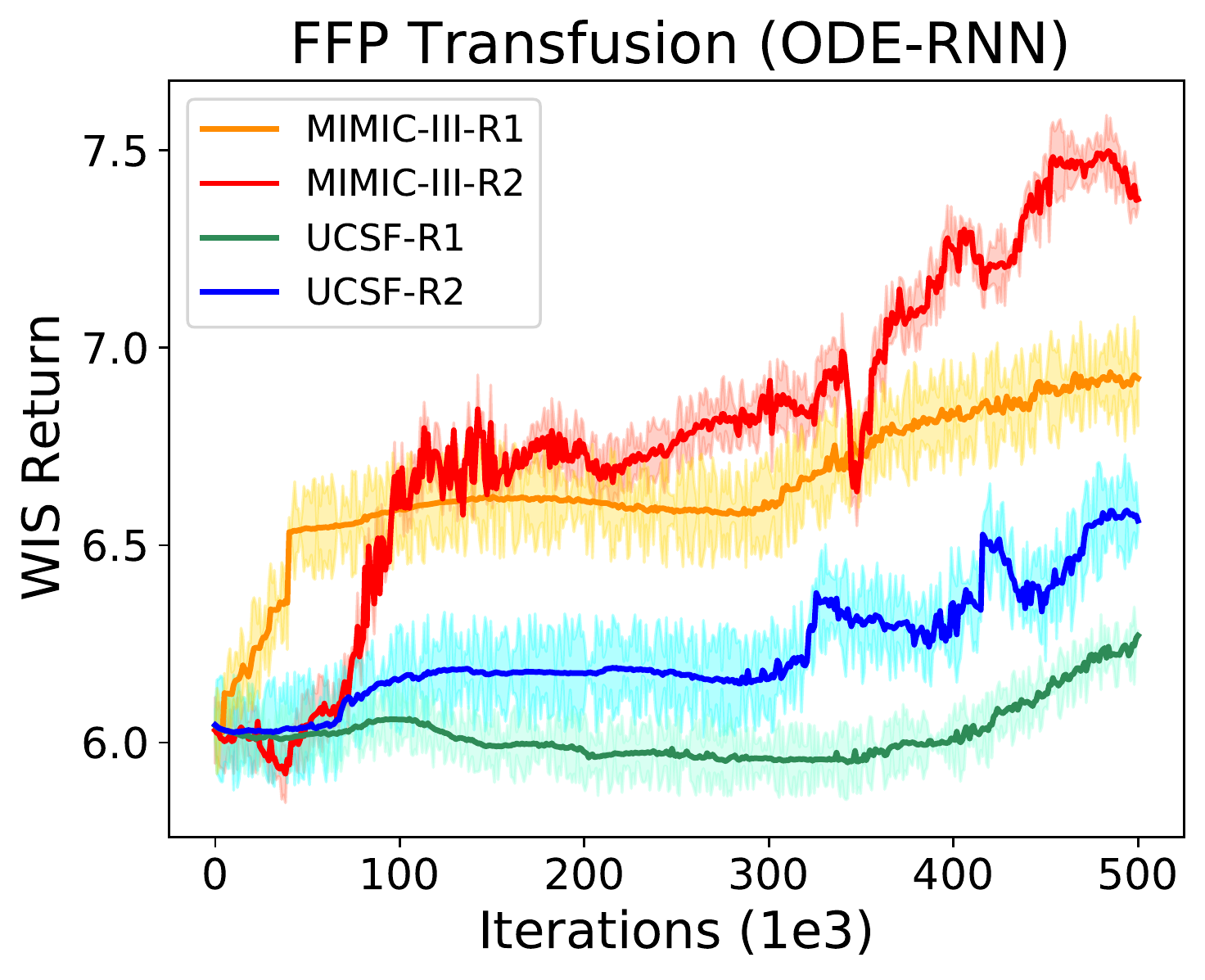}}
    \hfill
    {\includegraphics[width=0.33\textwidth, ]{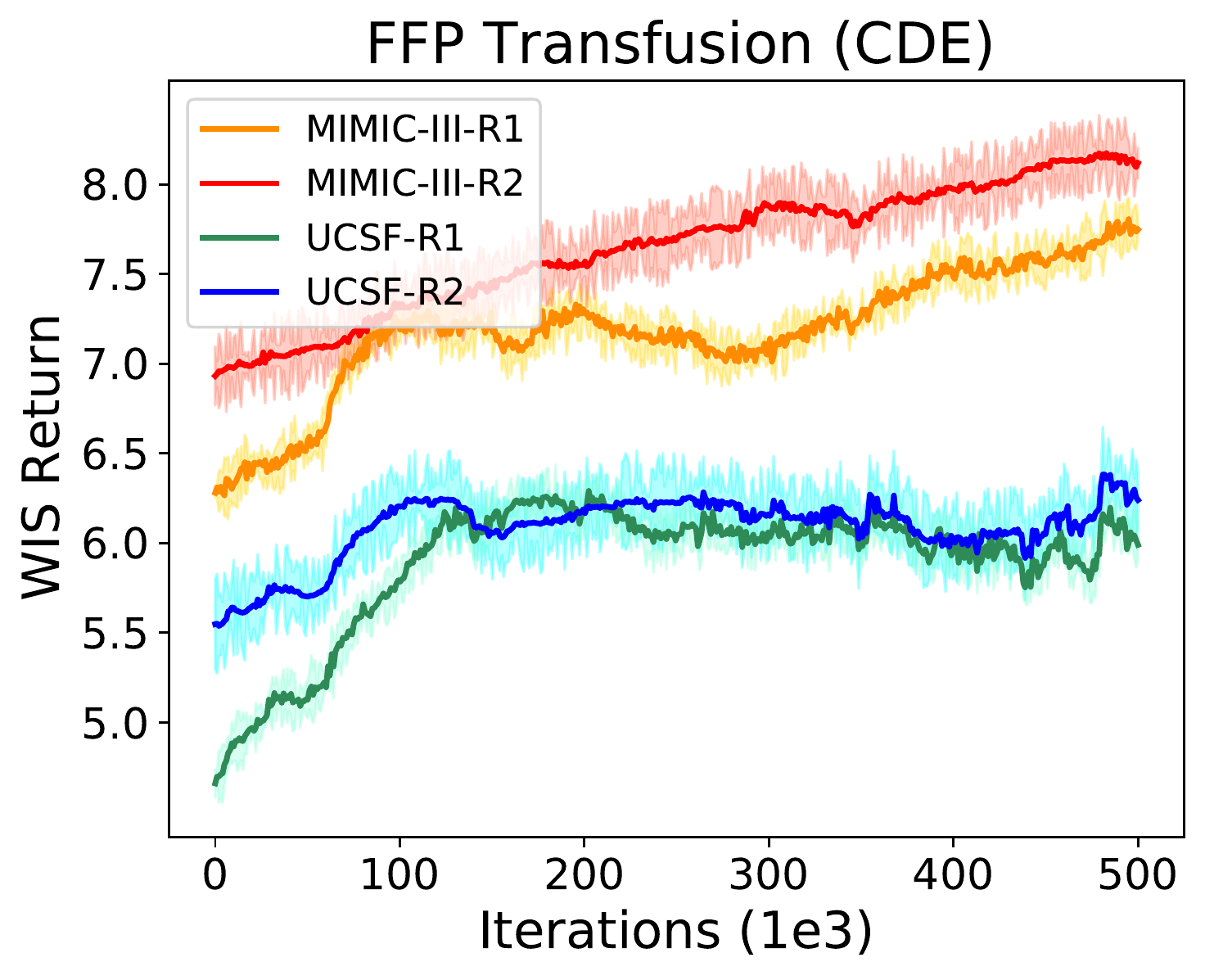}}
  \caption{WIS evaluation of policies (learning curves) with $3$ representation approaches (RNN, ODE-RNN, and CDE) and $2$ reward functions on three transfusion tasks. All policies are trained from a replay buffer comprised of the training batch of patient
trajectories for 500k iterations, evaluating the learned policy every 1000 iterations on the testing set of both datasets. The displayed results are averaged over $5$ random seeds. The shaded area measures a single standard deviation across seeds. Across all tasks, representations with RNN and CDE along with the R2 reward function generally have better policy learning curves compared to those using ODE-RNN and the R1 reward function. Furthermore, the policy learned on the MIMIC-III dataset is far more superior to the UCSF dataset according to the WIS return value and the stability of learning curves.}
  \label{fig:wis_eval}
\end{figure*}

\subsection{Experimental Settings}
\textbf{Training Infrastructure}: All of the experiments regarding state representations and policy learning were implemented in Pytorch on one NVIDIA Tesla P100 GPU.\\
\textbf{Data Splitting}: For both datasets, we randomly split the data in the ratio $70:15:15$. That is, 70\% data will go to the training set, 15\% to the validation set and remaining 15\% to the test set. \\
\textbf{Evaluation Metrics}: 
\begin{enumerate}
    \item[(I)]\textit{\textbf{Accuracy}}: When evaluating the closeness of a match between actions taken by RL agents and ground truth actions performed by physicians, we measured the performance in terms of accuracy. In this respect, we also consider a few classification models for comparison, which are commonly used in blood transfusion prediction tasks, including logistic regression (LR), random forest (RF), eXtreme Gradient Boosting (XGBOOST), and multi-layer perceptron (MLP). All experiments of these classification algorithms were implemented using scikit-learn.
    \item[(II)]\textit{\textbf{WIS}}: When evaluating off-policy learning from the RL perspective, we use WIS, a means to correct the mismatch between the probabilities of a trajectory under the behavior and target policies (learned policy using discretized BCQ), which is computed by: $R^{WIS} = \frac{\sum_{n=1}^N R_n w_n}{\sum_{n=1}^N w_n}$, where $w_n$ is the per-trajectory importance sampling weight, a fraction between the target policy $\pi$ and the behavior policy $\mu$,
and $R_n$ is the empirical outcome of trajectory $n$. Here, the behavior policy used in WIS was behavior cloning, a 2-layer fully connected (FC)
network with ReLU activation functions in between. It was trained with cross-entropy loss.
\end{enumerate}
\textbf{Architecture Details for Patient State Representations}:
\begin{enumerate}
    \item[(I)] \textit{\textbf{RNN}}: A 3-layer RNN is used to estimate the encoding function $\Phi_{\text{RNN}}$. The first two layers are FC layers with ReLU activation functions, followed by a gated recurrent unit layer. The dimension of the hidden state was selected by grid search among $\{32, 64, 128, 256\}$.
    \item[(II)] \textit{\textbf{ODE-RNN}}:  A gated recurrent unit with $100$ units is used to estimate the encoding function $\Phi_{\text{ODE-RNN}}$. The hidden states between feature observations are modeled by a neural ODE, parameterized by a 2-layer MLP with 100 hidden units. The adaptive step size is using the fifth-order \textit{dopri5} solver from the torchdiffeq package.  
    \item[(III)] \textit{\textbf{CDE}}: To estimate the encoding function $\Phi_{\text{CDE}}$, a neural CDE is used, parameterized by a MLP with three hidden layers, each of which has 100 hidden units. For the hidden layers, we use ReLU activation functions, and a tanh activation function for the final layer.
\end{enumerate} 
In order for efficient transfer of discretized BCQ from the MIMIC-III to the UCSF, the output vector dimension is the same when we construct patient representations on both datasets. \\ 
\textbf{Policy Training}: The discretized BCQ algorithm is used to train the policies. In our BCQ, the Q-network is a 3-layer FC network. We used a uniformly sampled replay buffer for training, which comprised of the training batch of patient trajectories for 500k iterations. Then, we evaluated the learned policy every 1000 iterations using the testing subset of the data. \\
\textbf{Hyperparameters}:
The hyperparameter search spaces for classification algorithms and discretized BCQ policy training are listed in Table~\ref{tab:classification_hyper} and Table~\ref{tab:bcq_hyper}, respectively. All three transfusion tasks use the same set of search space on both datasets. We perform grid search for hyperparameter optimization. 

\subsection{Off-Policy Evaluation}\label{ope}
We evaluate the learned policy from the discretized BCQ algorithm with WIS. Specifically, for each transfusion task, we consider the influence of state representations and reward mechanisms on policy learning. Figure~\ref{fig:wis_eval} presents all combinations of $3$ state presentations and $2$ reward mechanisms on both datasets for each task. Based on our reward design, the magnitude of the WIS value should lie in the $\pm 10$ possible range. Suppose that the behavior policy is close to the actual hospital policy, which can be considered as “experts” most of the time. Then, a higher WIS value corresponds to more closeness between the learned policy and the behavior policy, indicating that the learned policy is more effective. From Figure~\ref{fig:wis_eval}, it is evident that regardless of the RL setting (i.e., different encoders and rewards), the learned policies on the MIMIC-III dataset far outperform the ones learned on the UCSF dataset, primarily due to differences in batch sample sizes. Furthermore, the performance of transfusion policies is generally superior when using R2 mechanism to the ones using R1 mechanism based on the policy learning curves in Figure~\ref{fig:wis_eval}. Finally, the learned policies from patient state representations encoded by RNN and CDE outperform ODE-RNN. The learning curves of RNN and CDE demonstrate an overall steady growth with some small oscillations. For ODE-RNN, however, the learning curves show more volatility. Overall, oscillations and drops in performance during the intermediate iterations signify the impacts of encoded state representations, reward design mechanisms, and batch sample sizes on the policy learning of offline RL agents.

\begin{table*}[htbp]
  \caption{Accuracy comparison on the testing subsets between actions taken by RL agents / classification models and ground truth actions implemented by the hospital. Experiments are conducted with $5$ random initializations. The results are shown in the format of mean and standard deviation. Note that RNN-R1 represents the setting using the RNN state representation with reward function R1. Symbols for other settings in the descretized BCQ algorithm are similar.}
  \label{tab:accuracy_comp}
  \begin{tabular}{c c c c c c c c} 
\hline
& & \multicolumn{2}{c}{\textit{\textbf{RBC transfusion}}} & \multicolumn{2}{c}{\textit{\textbf{PLT transfusion}}} & \multicolumn{2}{c}{\textit{\textbf{FFP transfusion}}}  \\
\cmidrule(lr){3-4}\cmidrule(lr){5-6}\cmidrule(lr){7-8}
& & MIMIC-III & UCSF &  MIMIC-III & UCSF & MIMIC-III & UCSF \\ 
\hline
\multirow{6}{*} {\makecell{Discretized BCQ \\ Algorithm}} & RNN-R1 & 0.82 $\pm$ 0.02 & 0.67 $\pm$ 0.03 & 0.89 $\pm$ 0.03 & 0.70 $\pm$ 0.02 & 0.85 $\pm$ 0.01 & 0.71 $\pm$ 0.02\\
    & RNN-R2 & 0.84 $\pm$ 0.01 & 0.68 $\pm$ 0.02 & 0.90 $\pm$ 0.02 & 0.73 $\pm$ 0.03 & 0.90 $\pm$ 0.02 & 0.72 $\pm$ 0.03 \\
    & ODE-RNN-R1 & 0.80 $\pm$ 0.03 & 0.63 $\pm$ 0.02 & 0.86 $\pm$ 0.02 & 0.68 $\pm$ 0.02 & 0.87 $\pm$ 0.01 & 0.70 $\pm$ 0.02 \\
    & ODE-RNN-R2 & 0.81 $\pm$ 0.02 & 0.63 $\pm$ 0.01 & 0.87 $\pm$ 0.01 & 0.70 $\pm$ 0.02 & 0.86 $\pm$ 0.01 & 0.70 $\pm$ 0.02 \\
    & CDE-R1 & 0.84 $\pm$ 0.02 & 0.69 $\pm$ 0.02 & 0.89 $\pm$ 0.02 & 0.72 $\pm$ 0.01 & 0.83 $\pm$ 0.02 & 0.71 $\pm$ 0.02 \\
    & CDE-R2 & \textbf{0.85 $\pm$ 0.02} & 0.71 $\pm$ 0.01 & 0.89 $\pm$ 0.01 & 0.73 $\pm$ 0.02 & 0.90 $\pm$ 0.02 & 0.71 $\pm$ 0.01 \\
\hline
\multirow{4}{*} {\makecell{Classification \\ Algorithms}} & LR & 0.73 $\pm$ 0.02 & 0.69 $\pm$ 0.01 & 0.76 $\pm$ 0.02 & 0.75 $\pm$ 0.02 & 0.79 $\pm$ 0.01 & 0.73 $\pm$ 0.01 \\
    & RF & 0.84 $\pm$ 0.01 & 0.80 $\pm$ 0.02 & 0.89 $\pm$ 0.01 & 0.87 $\pm$ 0.01 & 0.91 $\pm$ 0.01 & \textbf{0.84 $\pm$ 0.02}\\
    & XGBOOST & \textbf{0.85 $\pm$ 0.02} & \textbf{0.82 $\pm$ 0.01} & \textbf{0.92 $\pm$ 0.01} & 0.88 $\pm$ 0.02 & \textbf{0.93 $\pm$ 0.02} & 0.83 $\pm$ 0.01 \\
    & MLP & 0.82 $\pm$ 0.03 & 0.79 $\pm$ 0.02 & 0.89 $\pm$ 0.01 & \textbf{0.89 $\pm$ 0.02} & 0.90 $\pm$ 0.02 & 0.82 $\pm$ 0.01\\
\hline
\end{tabular}
\end{table*}

\subsection{Degree of Matching between RL agents and Physicians’ Decisions}\label{matching}
In addition to evaluating the learned policy using standard RL off-policy evaluation via WIS, we further compare the discrete BCQ algorithm's recommendations with respect to blood transfusion against the true policy implemented by the hospital. In addition, $4$ classification models (LR, RF, XGBOOST, and MLP) are considered for accuracy comparison. From the RL perspective, in consideration of the performance influence using different state representations and reward designs, we consider all combinations of representation approaches and reward design mechanisms and evaluate their performance. We report all the results on three transfusion tasks, as summarized in Table~\ref{tab:accuracy_comp}. For the MIMIC-III dataset, some settings of the discretized BCQ algorithm (RNN-R2 and CDE-R2) regarding action recommendations achieved comparable performance to most classification algorithms. Comparing performances across all RL settings in terms of state representations and rewards, policies learned from representations encoded by RNN and CDE outperform those using ODE-RNN, indicating that the representations from ODE-RNN did not adequately encode sufficient information to learn a policy from the batch mode, perhaps due to limited data. Furthermore, using R2 mechanism generally performs better than using R1 mechanism. This may be primarily due to the fact that R2 mechanism is more clinically guided such that it may reflect patients' real-time condition change better than using R1 mechanism. For the UCSF dataset, the accuracy performance of RL agents far underperform classification models regardless of the RL setting, mainly due to its tiny data size, making the RL agent more difficult to learn. A potential solution using knowledge from external expertise such as the MIMIC-III to overcome this issue will be discussed in Section~\ref{transfer}.

\subsection{Transfer RL}\label{transfer}
\begin{figure}[htbp]
    \centering
    {\includegraphics[width=0.235\textwidth,]{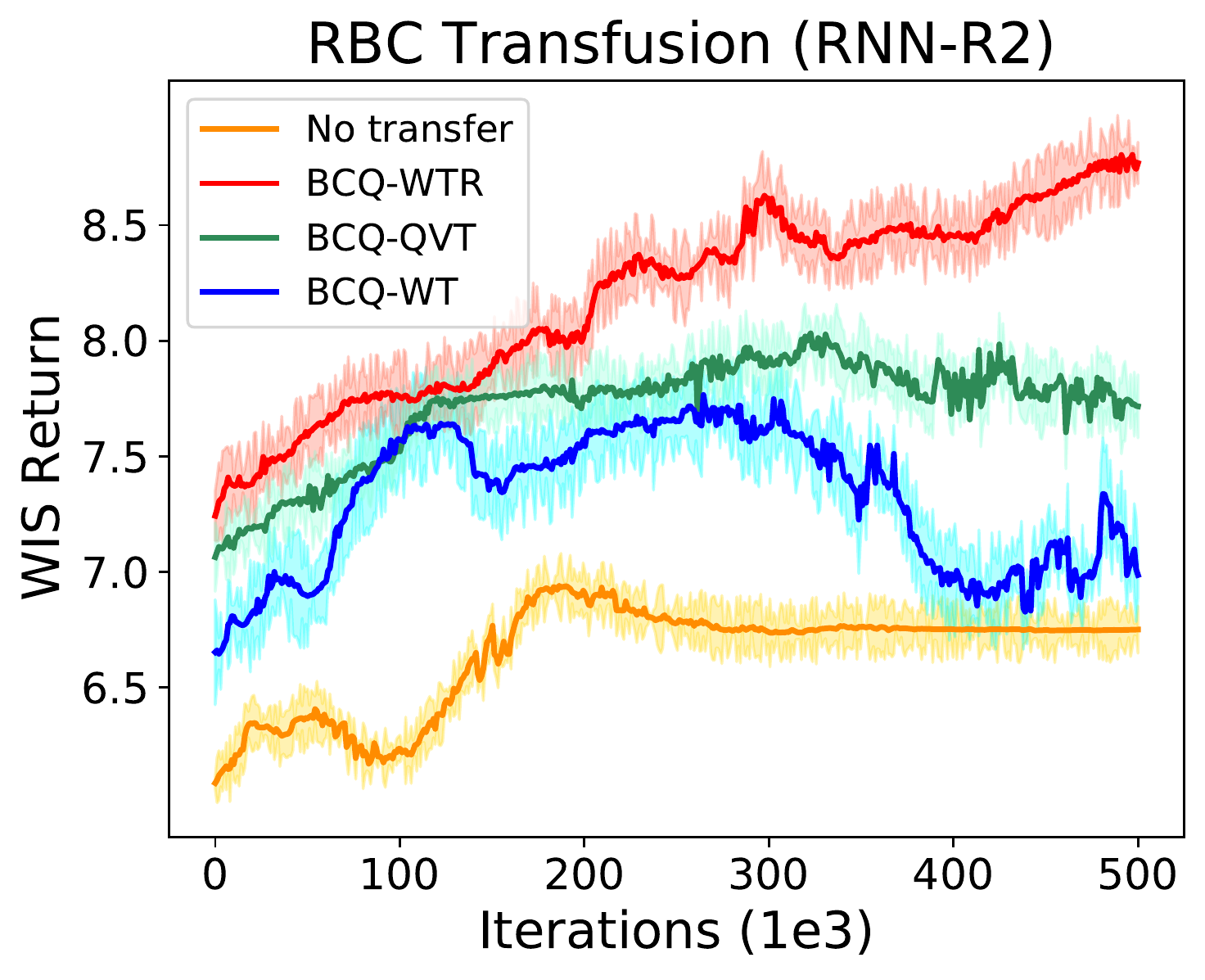}}
    \hfill
    {\includegraphics[width=0.235\textwidth, ]{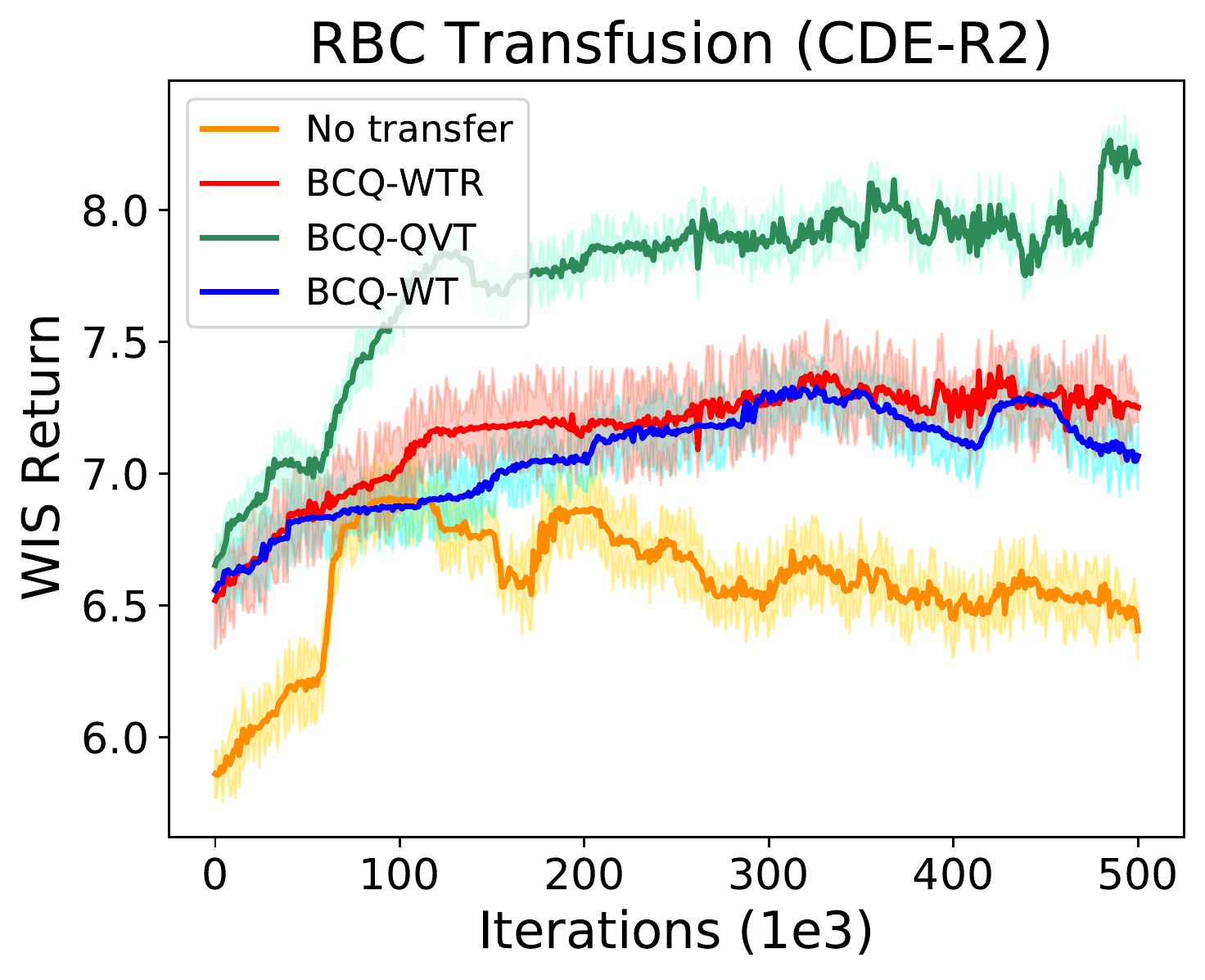}}
    \hfill
    {\includegraphics[width=0.235\textwidth, ]{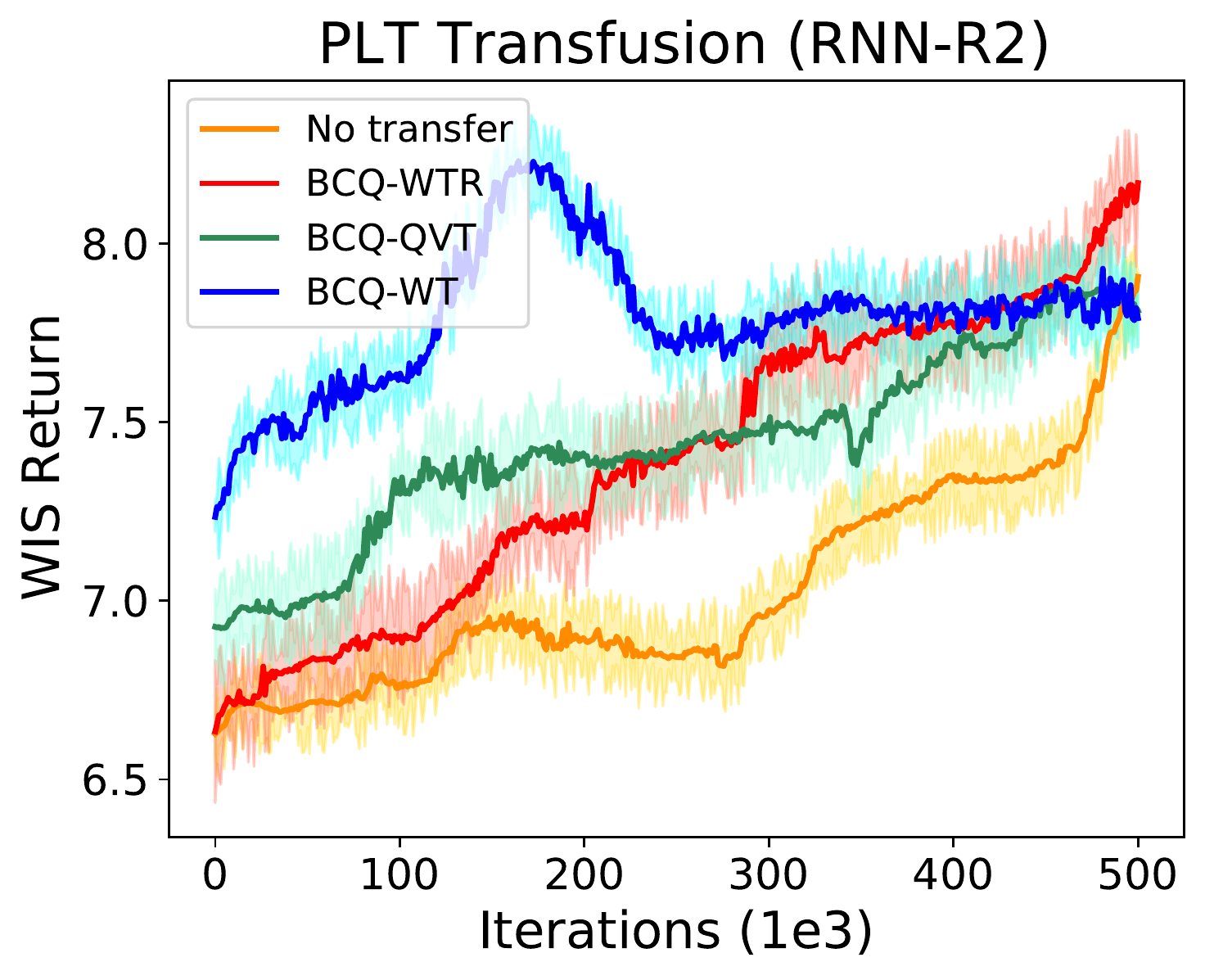}}
    \hfill
    {\includegraphics[width=0.235\textwidth,]{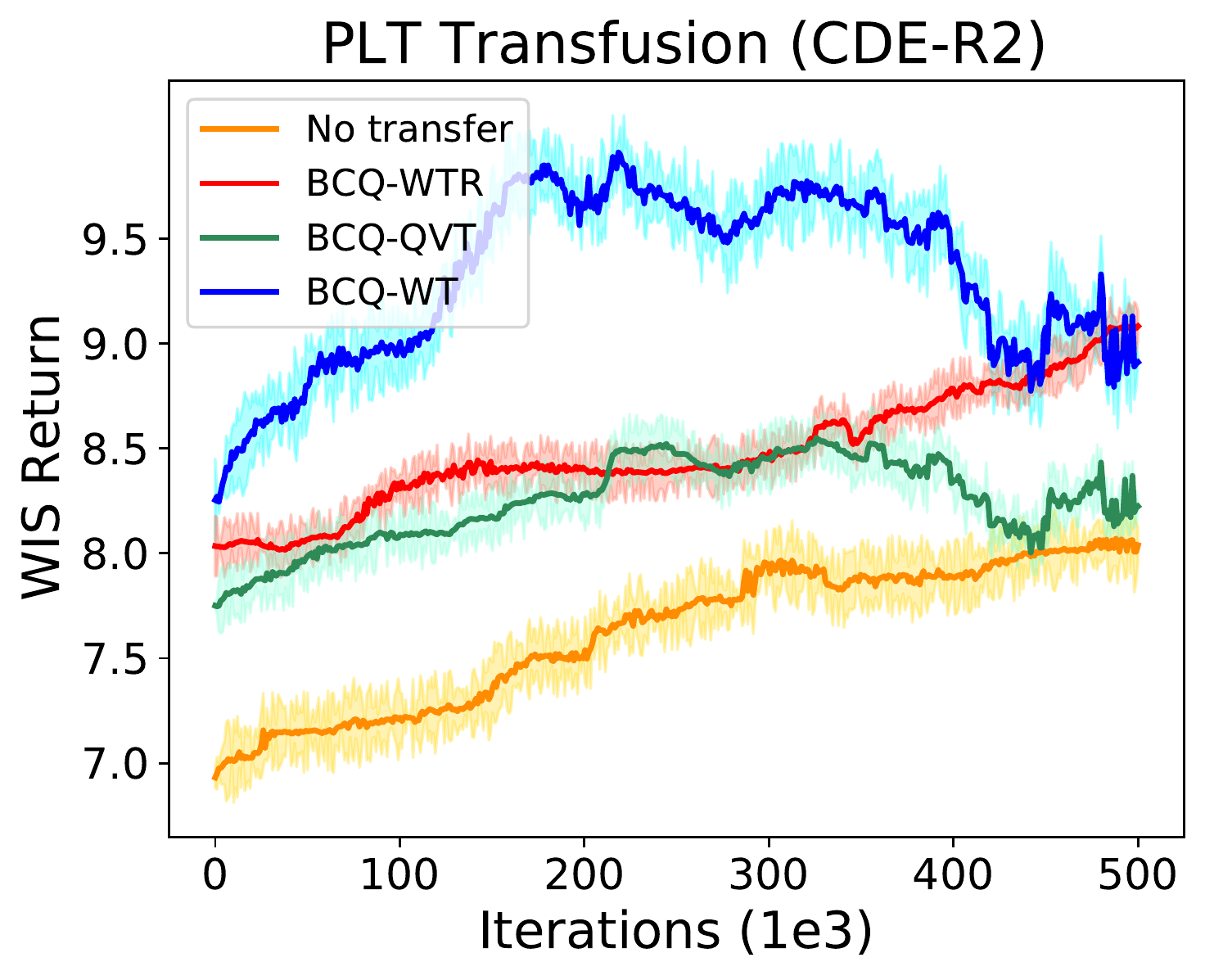}}
    \hfill
    {\includegraphics[width=0.235\textwidth, ]{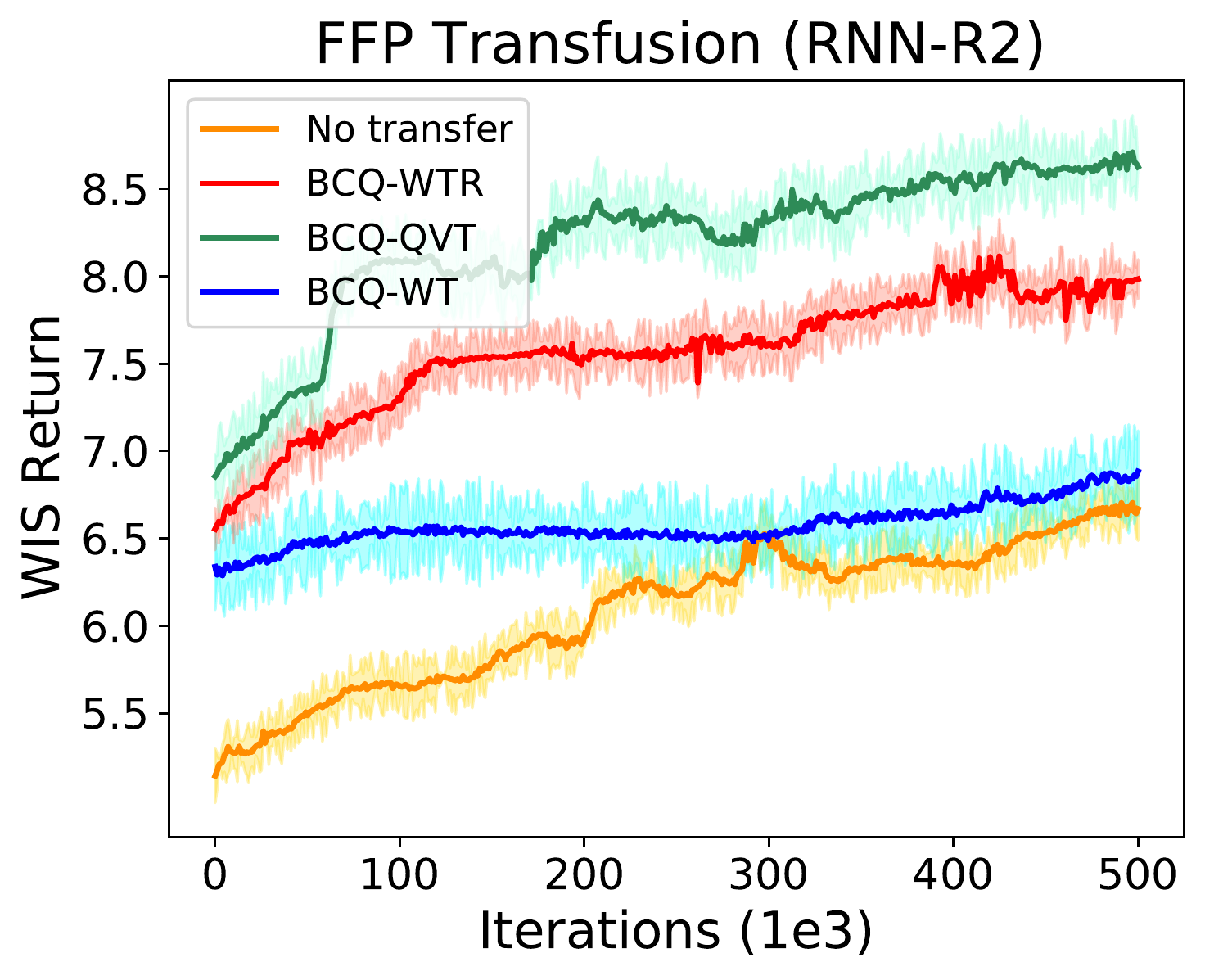}}
    \hfill
    {\includegraphics[width=0.235\textwidth, ]{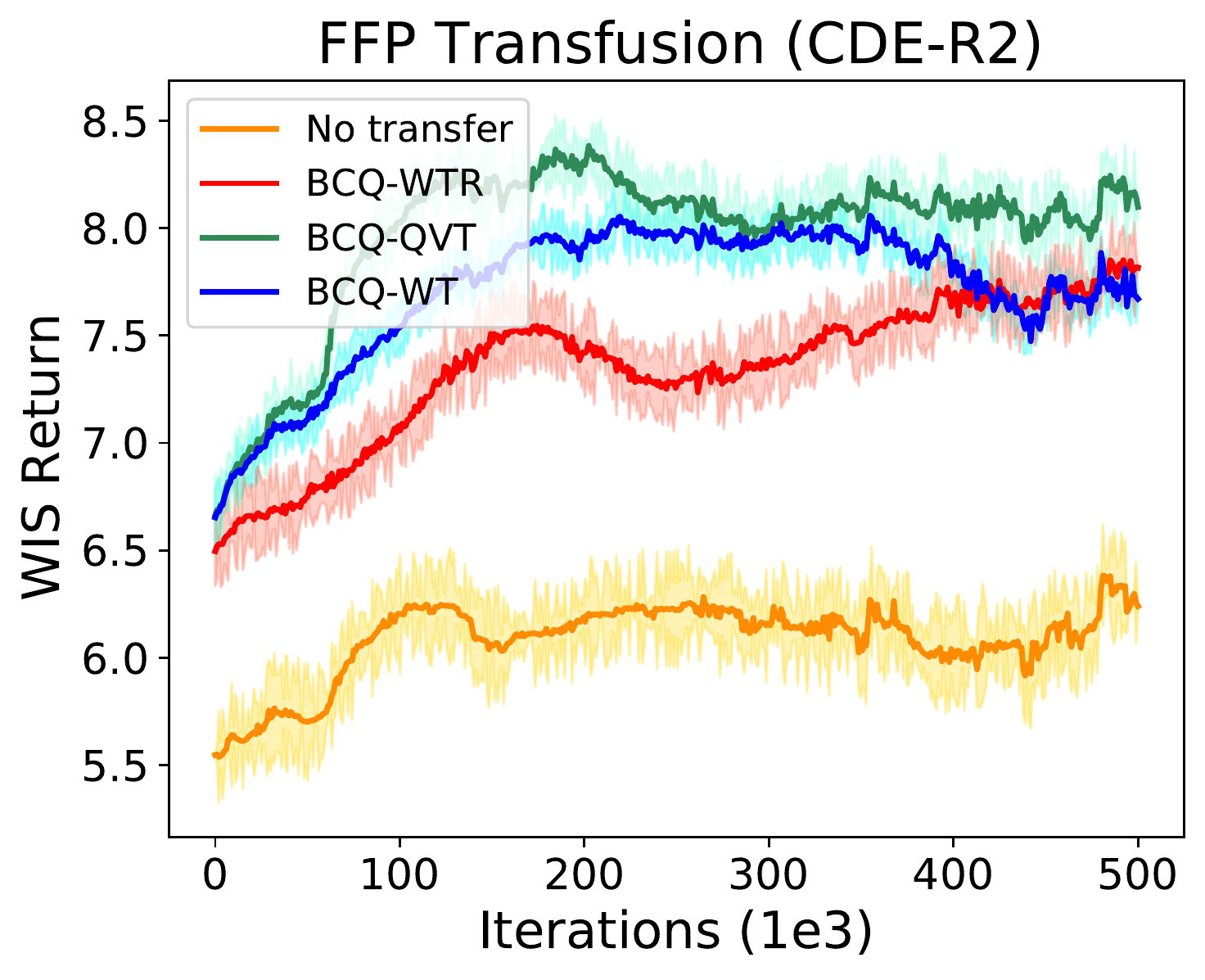}}
  \caption{WIS evaluation of policies (learning curves) on the UCSF testing subset with RNN-R2 and CDE-R2 RL settings with and without TL on three transfusion tasks. The displayed results are averaged over $5$ random seeds. The shaded area measures a single standard deviation across seeds. Among three tasks, all transfer methods show better jump-start and asymptotic performance compared to the original policy learning curves without TL. For RL settings like RNN-R2, transfer can help reduce the oscillations in FFP and PLT transfusion tasks.}
  \label{fig:wis_eval_transfer}
\end{figure}

Due to the poor performance of policy learning on the UCSF dataset, we use external knowledge from MIMIC-III to improve its performances in terms of WIS and accuracy. As discussed in Section~\ref{BCQ_transfer}, we consider three types of transfer and evaluate their performance: BCQ-QVT, BCQ-WT (retraining all layers without re-initializing any layers), and BCQ-WTR (re-initializing the FC layers and retraining all layers). Based on the results from Section~\ref{ope} and Section~\ref{matching}, we only consider the RL settings using R2 mechanism with state representations RNN and CDE. Results of policy learning on the UCSF dataset with and without TL are presented in Figure~\ref{fig:wis_eval_transfer} and Table~\ref{tab:accuracy_transfer}. All three transfer methods yielded better performance than the original policy learning on the UCSF dataset in terms of WIS and accuracy. Specifically, from Figure~\ref{fig:wis_eval_transfer}, all transfer methods show better jump-start performance (i.e., the initial performance of the agent) and asymptotic performance (i.e., the ultimate performance of the agent) compared to the original policy learning curves without TL. In particular, across all three tasks, the jump-start WIS return improves up to $18.94\%$ and the asymptotic WIS return improves up to $21.63\%$ on average. For some transfusion tasks such as FFP transfusion with RNN-R2 and PLT transfusion with RNN-R2, some transfer methods can help reduce the oscillations, resulting in more steady growth learning curves. For some other tasks, transfer methods like BCQ-WT even bring about larger oscillations, indicating more unstable policy learning compared to the performances without TL. This may be due to weights of different scales between two datasets and pre-trained policies from the MIMIC-III dataset may get stuck in local optima. In Table~\ref{tab:accuracy_transfer}, among three transfusion tasks, the degree of matching between the transferred policy and the ground truth policy has significantly improved to different extents. On average, the accuracy improves up to $17.02\%$. Generally, both evaluation metrics demonstrate the effectiveness of TL on offline agent policy learning.

\begin{table*}[htbp]
  \caption{Accuracy comparison on the UCSF testing subset between actions taken by RL agents ground truth actions implemented by the hospital with and without TL. Experiments are conducted with $5$ random initializations. The results are shown in the format of mean and standard deviation. The accuracy improves to varying extents after transfer.}
  \label{tab:accuracy_transfer}
  \begin{tabular}{c| c c c c c c c c} 
\hline
 & \makecell{No Transfer \\ (RNN-R2)} &  \makecell{No Transfer \\ (CDE-R2)} & \makecell{BCQ-QVT \\ (RNN-R2)} & \makecell{BCQ-QVT \\ (CDE-R2)} & \makecell{BCQ-WT \\ (RNN-R2)} & \makecell{BCQ-WT \\ (CDE-R2)} & \makecell{BCQ-WTR \\ (RNN-R2)} & \makecell{BCQ-WTR \\ (CDE-R2)} \\
 \hline
 \textit{\textbf{RBC transfusion}} & 0.68 $\pm$ 0.02 & 0.71 $\pm$ 0.01 & 0.78 $\pm$ 0.01 & \textbf{0.81 $\pm$ 0.02} & 0.74 $\pm$ 0.03 & 0.75 $\pm$ 0.02 & 0.79 $\pm$ 0.02 & 0.80 $\pm$ 0.02\\
 \textit{\textbf{PLT transfusion}} & 0.73 $\pm$ 0.03 & 0.73 $\pm$ 0.02 & 0.82 $\pm$ 0.02 & 0.81 $\pm$ 0.01 & 0.78 $\pm$ 0.03 & 0.78 $\pm$ 0.02 & \textbf{0.85 $\pm$ 0.02} & \textbf{0.85 $\pm$ 0.02}\\
 \textit{\textbf{FFP transfusion}} & 0.72 $\pm$ 0.03 & 0.71 $\pm$ 0.01 & 0.79 $\pm$ 0.01 & \textbf{0.82 $\pm$ 0.02} & 0.78 $\pm$ 0.03 & 0.79 $\pm$ 0.02 & \textbf{0.82 $\pm$ 0.01} & \textbf{0.82 $\pm$ 0.02} \\
\hline
\end{tabular}
\end{table*}

\subsection{Policy Simulation}
All the above analysis are based on the assumption that the actual physician decision making regarding blood transfusion can improve patients' clinical outcomes. However, in practice, physicians may not be able to always make optimal transfusion decisions due to insufficient communication with patients and an incomplete understanding of patients' conditions and medical history due to urgency in the ICU. Furthermore, transfusion may not always improve patients' outcomes, resulting in higher risk of mortality and morbidity. Here, we consider patients' short-term (decreased acuity scores) and long-term (increased survival rates) clinical outcomes. In our two real-world datasets, we calculate the Pearson correlation coefficients between a decision of giving transfusion and clinical outcomes on all tasks, as shown in Table~\ref{tab:Pearson}. 
\begin{table}[ht]
  \caption{Pearson correlation between transfusion decision and clinical outcomes. LT  and ST are short for long-term and short-term, respectively.}
  \label{tab:Pearson}
  \begin{tabular}{ccc}

    \toprule
    Intervention & \makecell{MIMIC-III \\ LT / ST} &  \makecell{UCSF \\ LT / ST}\\
    \midrule
    RBC transfusion & 0.35 / 0.29 & -0.21 / 0.08 \\
    PLT transfusion & 0.38 / 0.23 & -0.25 / -0.12 \\
    FFP transfusion & 0.32 / 0.31 & -0.33 / -0.19 \\
  \bottomrule
\end{tabular}
\end{table}
For the MIMIC-III dataset, transfusion does improve patients' both long-term and short-term outcomes. For the UCSF dataset, however, the negative coefficients indicate that not all transfusion decisions improve patients' conditions. Hence, actual physicians' real-time treatment strategies of blood transfusion may require further optimization, probably by assistance from RL agents. Hence, we extract patients from the UCSF dataset that receive transfusions during their hospital stay but decease within 28 days after their admission. Then, we perform policy simulations from the transferred transfusion policies on the UCSF dataset to model the environment of real-time transfusion decision changes and corresponding patients' outcomes. Here, we use the MIMIC-III dataset to help with the simulation process. We group patients from the MIMIC-III dataset and the selected patients from the UCSF dataset by clustering patients according to their temporal value changes and static demographics. Then, by using patients from the same cluster on the MIMIC-III dataset who survived in 28 days after ICU admission as a control group, we conducted simulations by changing the real-time transfusion policy on the selected UCSF patient cohorts. The estimated mortality rate over all UCSF patients decreases from the actual 16.48$\%$ to 13.74$\%$. Similarly, we select patients with worsening conditions during their hospital stay on the UCSF dataset and conduct simulations via transferred policies. The estimated decreased acuity rate over all UCSF patients decreases from the actual 9.13$\%$ to 7.95$\%$. Here, all the results are averaged over three transfusion tasks. This is an important finding which supports decision making tools have significant potential to improve patient outcomes.

\section{Conclusions and Future Work}\label{conclusion}
In this work, we utilized an off-policy batch reinforcement learning algorithm, discretized BCQ, to tackle policy recommendations for blood transfusion in ICUs. We conduct experiments on two real-world datasets with different patient state encoding and reward function mechanisms. Our results demonstrate that using appropriate state representations like RNN and CDE, along with proper reward designs like R2 can provide reasonably well policy training. Furthermore, an integration of TL and RL can help improve the policy learning on a data-scarce dataset to a large extent. As a decision support tool, the learned policy by RL agents may serve as an auxiliary advice for physicians in emergency, and thus potentially assist physicians to optimize the real-time treatment strategies on blood transfusion. Hence, blending the RL with real physicians' decisions using available patient information could lead to better transfusion strategies and improving patients' clinical outcomes. Possible directions for future work include exploring different patient state representations for better policy learning, extending the action space to include continuous transfusion dosages, using more principled approach to the design of the rewards such as inverse RL, adopting different evaluation methods like doubly robust evaluation, as well as applying the method to other unexplored clinical decision making problems that may fit the RL setting. 

\section{Acknowledgments}
This work was funded by the National Institutes for Health (NIH) grant
NIH 7R01HL149670. We greatly thank Dr. Rachael A. Callcut
from the University of California, Davis with essential discussions.

\bibliographystyle{ACM-Reference-Format}
\bibliography{main}


\end{document}